\documentclass[10pt,journal,compsoc]{IEEEtran}
\ifCLASSOPTIONcompsoc
  \usepackage[nocompress]{cite}
\else
  \usepackage{cite}
\fi
\ifCLASSINFOpdf
\else
\fi

\usepackage{graphicx}
\usepackage{amsmath}
\usepackage{amssymb}
\usepackage{color}
\usepackage{graphics}
\usepackage{multirow}
\usepackage{subfig}
\usepackage{array}
\usepackage{colortbl}
\usepackage{cite}
\usepackage{diagbox}
\usepackage{rotating}
\usepackage{booktabs}
\usepackage{overpic}
\usepackage{contour}
\usepackage{ragged2e}
\usepackage[pagebackref=false,breaklinks=true,letterpaper=true,colorlinks,citecolor=black,bookmarks=false]{hyperref}

\def\ie{\emph{i.e.}}
\def\eg{\emph{e.g.}}

\newcommand{\fdp}[1]{\textcolor{black}{#1}}
\newcommand{\fkr}[1]{\textcolor{black}{#1}}

\def\ourmodel{\emph{JL-DCF}}
\graphicspath{{./Imgs/}}
\newcommand{\tabincell}[2]{\begin{tabular}{@{}#1@{}}#2\end{tabular}}
\begin{document}
%
\title{Siamese Network for RGB-D\\ Salient Object Detection and Beyond}
%
%
%
%

\author{Keren Fu,~ 
Deng-Ping Fan$^*$,~
Ge-Peng Ji,~
Qijun Zhao,~ 
Jianbing Shen,~
and Ce Zhu, \IEEEmembership{Fellow,~IEEE}
\IEEEcompsocitemizethanks{
\IEEEcompsocthanksitem Keren Fu and Qijun Zhao are with the College of Computer Science, Sichuan University, China, and are also with the National Key Laboratory of Fundamental Science on Synthetic Vision, Sichuan University. (Email: fkrsuper@scu.edu.cn, qjzhao@scu.edu.cn)
\IEEEcompsocthanksitem Deng-Ping Fan is with the College of Computer Science, Nankai University, Tianjin, China. (Email: dengpingfan@mail.nankai.edu.cn)
\IEEEcompsocthanksitem Ge-Peng Ji is with the School of Computer Science, Wuhan University, China. (Email: gepengai.ji@gmail.com)
\IEEEcompsocthanksitem Jianbing Shen is with the Inception Institute of Artificial Intelligence (IIAI), United Arab Emirates.
(Email: shenjianbingcg@gmail.com)
\IEEEcompsocthanksitem Ce Zhu is with the School of Information and Communication Engineering, University of Electronic Science and Technology of China. (Email: eczhu@uestc.edu.cn)
\IEEEcompsocthanksitem A preliminary version of this work has appeared in CVPR 2020~\cite{fu2020jldcf}.
\IEEEcompsocthanksitem Corresponding author: Deng-Ping Fan.
}
}

%
%

\markboth{Journal of \LaTeX\ Class Files,~Vol.~14, No.~8, August~2015}%
{Shell \MakeLowercase{\textit{et al.}}: Bare Demo of IEEEtran.cls for Computer Society Journals}
%



\IEEEtitleabstractindextext{%
\begin{abstract}
\justifying
   Existing RGB-D salient object detection (SOD) models usually treat RGB and depth as independent information and design separate networks for feature extraction from each. Such schemes can easily be constrained by a limited amount of training data or over-reliance on an elaborately designed training process. Inspired by the observation that RGB and depth modalities actually present certain commonality in distinguishing salient objects, a novel joint learning and densely cooperative fusion (\emph{\ourmodel}) architecture is designed to learn from both RGB and depth inputs through a shared network backbone, known as the \emph{Siamese architecture}. In this paper, we propose two effective components: joint learning (JL), and densely cooperative fusion (DCF). The JL module provides robust saliency feature learning by exploiting cross-modal commonality via a Siamese network, while the DCF module
   is introduced for complementary feature discovery. Comprehensive experiments using five popular metrics show that the designed framework yields a robust RGB-D saliency detector with good generalization. As a result, JL-DCF significantly advances the state-of-the-art models by an average of $\sim$2.0\% (max F-measure) across seven challenging datasets. In addition, we show that \emph{\ourmodel}~is readily applicable to other related multi-modal detection tasks, including RGB-T (thermal infrared) SOD and video SOD, achieving comparable or even better performance against state-of-the-art methods. \fkr{We also link \emph{\ourmodel}~to the RGB-D semantic segmentation field, showing its capability of outperforming several semantic segmentation models on the task of RGB-D SOD.} These facts further confirm that the proposed framework could offer a potential solution for various applications and provide more insight into the cross-modal complementarity task.
\end{abstract}


}

\maketitle

\IEEEdisplaynontitleabstractindextext

%
\IEEEpeerreviewmaketitle

\IEEEraisesectionheading{\section{Introduction}}

\IEEEPARstart{S}{alient} object detection (SOD) aims at detecting the objects in a scene that humans would naturally focus on \cite{cheng2015global,borji2014salient,zhao2019egnet}. It has numerous useful applications, including
object segmentation/recognition \cite{Liu2012,ye2017salient,zhoucvpr2020,jerripothula2016image,
Rutishauser2004,han2006unsupervised}, image/video compression \cite{Guo2010},
video detection/summarization \cite{Ma2005,fan2019shifting}, content-based image editing
\cite{wang2017deep,Stentiford2007,Marchesotti2009,Ding2011,Goferman2010}, informative common object
discovery \cite{zhang2016detection,zhang2017co,zhang2020adaptive}, and image retrieval \cite{Chen2009,Gao2015,liu2013model}.
Many SOD models have been developed under the assumption that the inputs are individual
RGB/color images \cite{wang2016correspondence,zhang2017amulet,zhang2017learning,zhang2018progressive,feng2019attentive,piao2019deep,fu2019deepside} or sequences \cite{wang2019learning,wang2019zero,song2018pyramid,wang2019revisiting,fang2019salient}. As depth cameras, such as Kinect and RealSense, become more and more
popular, SOD from RGB-D inputs (``D'' refers to depth) is emerging as an attractive research
topic. Although a number of previous works have attempted to explore the role of depth
in saliency analysis, several issues remain:

\begin{figure}
\includegraphics[width=0.485\textwidth]{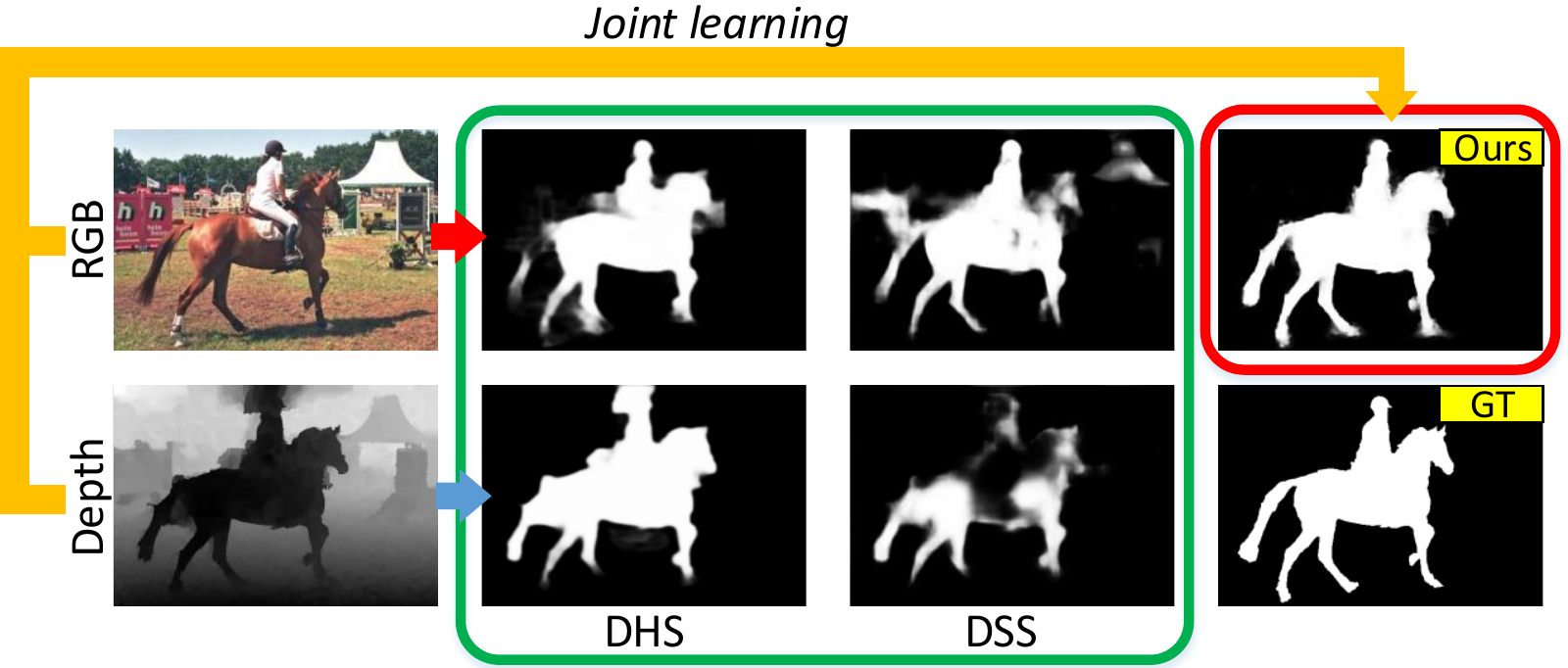}
\vspace{-15pt}
\caption{Applying deep saliency models DHS \cite{liu2016dhsnet} and
DSS \cite{hou2019deeply}, which are fed with an RGB image (1$^{st}$ row) or a depth map (2$^{nd}$ row).
Both of the models are trained on a single RGB modality. By contrast, our \ourmodel~considers both modalities and thus generates better results (last column).}
\label{fig_motivation}

\end{figure}

\textbf{(i) Deep-based RGB-D SOD methods are still under-explored:}
Despite more than one hundred papers on RGB SOD models being published since 2015 \cite{Fan2018SOC,wang2019salient,wang2019iterative,wang2018salient,wang2019salient2}, there are only a few deep
learning-based works focusing on RGB-D SOD. The first model utilizing
convolutional neural networks (CNNs) for RGB-D SOD \cite{qu2017rgbd}, which adopts a shallow CNN as the saliency map integration model, was introduced in 2017. Since then, only a dozen deep models have been proposed, as summarized in \cite{fan2019rethinking,Zhang2020UCNet}, leaving large room for further improvement in performance.

\textbf{(ii) Ineffective feature extraction and fusion:}
Most learning-based models fuse features of different modalities
either by early-fusion \cite{song2017depth,liu2019salient,huang2018rgbd,fan2019rethinking,zhang2021bilateral} or late-fusion \cite{han2017cnns,wang2019adaptive}.
Although these two simple strategies have achieved encouraging
progress in this field in the past (as pointed out in \cite{chen2018progressively}), they face challenges in either extracting representative multi-modal
features or effectively fusing them. While some other works have adopted a middle-fusion strategy \cite{chen2018progressively,zhu2019pdnet,chen2019three},
which conducts independent feature extraction and fusion using individual CNNs, their
sophisticated network architectures and large number of parameters require an elaborately designed training process and large amount of training data.
Unfortunately, high-quality depth maps are still sparse \cite{zhao2019contrast},
which may lead to sub-optimal solutions of deep learning-based models.

\noindent\textbf{Motivation.} To tackle RGB-D SOD, we propose a novel joint learning and densely cooperative
fusion (\ourmodel) architecture that
outperforms existing deep learning-based techniques.
Our method adopts the middle-fusion strategy mentioned above.
However, different from previous works which conduct independent feature
extraction from RGB and depth views\footnote{In this paper, ``view'' and ``modality'' are used interchangeably.}, \ourmodel~effectively extracts deep
hierarchical features from both inputs simultaneously, through a Siamese network \cite{chopra2005learning} (shared backbone).
The underlying motivation is that, although depth and RGB
images come from different modalities, they nevertheless share similar features/cues, such as strong figure-ground contrast \cite{niu2012leveraging,
peng2014rgbd,cheng2014depth}, closure of object contours \cite{feng2016local,
shigematsu2017learning}, and connectivity to image borders \cite{wang2017rgb,
liang2018stereoscopic}. This makes cross-modal transferring feasible, even for deep models. As evidenced in Fig. \ref{fig_motivation}, a model trained on a single RGB modality, like DHS \cite{liu2016dhsnet}, can sometimes perform well in the depth view. Nevertheless, a similar model, like DSS \cite{hou2019deeply}, could also fail in the depth view without proper adaption or transferring.

To the best of our knowledge, the proposed \ourmodel~scheme is \emph{the first to
leverage such transferability for deep models}, by treating a depth image as a
special case of a color image and employing a Siamese CNN for both RGB and depth
feature extraction. Additionally, we develop a densely cooperative fusion strategy to
reasonably combine the learned features of different modalities. In a nutshell, this paper provides three main contributions:

\begin{itemize}
\item This work is the first to leverage the commonality and transferability between RGB and depth views through a Siamese architecture. As a result, we introduce a general framework for RGB-D SOD,
called \ourmodel, which consists of two sub-modules: joint learning and
densely cooperative fusion. The key features of these two components are their robustness and effectiveness, which will be beneficial for future modeling in related multi-modality tasks in computer vision.
In particular, we advance the state-of-the-art (SOTA) by a significant average of $\sim$2\% (max F-measure) across seven challenging datasets. Further, by improving \ourmodel~through bridging between RGB and RGB-D SOD, even more gains can be obtained (see Section \ref{sec43}).
The code is available at
\href{https://github.com/kerenfu/JLDCF}{https://github.com/kerenfu/JLDCF/}.

\item We present a thorough evaluation of 14 SOTA methods~\cite{ju2014depth,feng2016local,cong2016saliency,song2017depth,
guo2016salient,qu2017rgbd,wang2019adaptive,han2017cnns,chen2019multi,
chen2018progressively,chen2019three,zhao2019contrast,fan2019rethinking,
Piao2019depth}. 
Besides, we conduct a comprehensive ablation study, including using different input sources, learning schemes, and feature fusion strategies, to demonstrate the effectiveness of \ourmodel.
Some interesting findings also encourage further research in this field.

\item In our experiments, we show that, in addition to the RGB-D SOD task, \ourmodel~is also directly applicable to other multi-modal detection tasks, including RGB-T (thermal infrared) SOD and video SOD (VSOD). Again, \ourmodel~achieves comparable or better performance against SOTA methods on these two tasks, further validating its robustness and generality. To the best of our knowledge, this appears to be the first time in the saliency detection community that a proposed framework is proved effective on such diverse tasks. \fkr{In addition, we conduct the first attempt to link \ourmodel~to RGB-D semantic segmentation, which is a closely related field, compare it with SOTA segmentation models, and draw underlying discussions.}
\end{itemize}

The remainder of the paper is organized as follows. Section \ref{sec2} discusses related work on RGB-D SOD, Siamese networks in computer vision, \fdp{and also RGB-D semantic segmentation}. Section \ref{sec3} describes the proposed \ourmodel~in detail. Experimental results, performance evaluations and comparisons are included in Section \ref{sec4}. Finally, conclusions are drawn in Section \ref{sec5}.
\section{Related Work}\label{sec2}

\subsection{RGB-D Salient Object Detection}\label{sec21}

\noindent\textbf{Traditional models.}
The pioneering work for RGB-D SOD was produced by Niu \emph{et al.} \cite{niu2012leveraging}, who introduced disparity contrast and domain knowledge into stereoscopic photography to measure stereo saliency. After Niu's work, various hand-crafted features/hypotheses originally proposed for RGB SOD were extended to RGB-D, such as center-surround difference\cite{ju2014depth,guo2016salient}, contrast \cite{cheng2014depth,peng2014rgbd,cong2016saliency}, background enclosure \cite{feng2016local}, center/boundary prior \cite{cheng2014depth,liang2018stereoscopic,wang2017rgb,cong2019going}, compactness \cite{cong2016saliency,cong2019going}, or a combination of various saliency measures \cite{song2017depth}. All the above models rely heavily on heuristic hand-crafted features, resulting in limited generalizability in complex scenarios.

\vspace{5pt}
\noindent\textbf{Deep models.}
Recent advances in this field have been obtained by using deep learning and CNNs.
Qu \emph{et al.} \cite{qu2017rgbd} first utilized a CNN to fuse different low-level saliency cues for judging the saliency confidence values of superpixels. Shigematsu \emph{et al.} \cite{shigematsu2017learning} extracted ten superpixel-based hand-crafted depth features capturing the background enclosure cue, depth contrast, and histogram distance. These features are fed to a CNN, whose output is shallowly fused with the RGB feature output to compute superpixel saliency.

A recent trend in this field is to exploit fully convolutional neural networks (FCNs) \cite{Long2017Fully}. Chen \emph{et al.} \cite{chen2018progressively} proposed a bottom-up/top-down architecture \cite{pinheiro2016learning}, which progressively performs cross-modal complementarity-aware fusion in its top-down pathway. Han \emph{et al.} \cite{han2017cnns} modified/extended the structure of the RGB-based deep neural network in order for it to be applicable for the depth view and then fused the deep representations of both views via a fully connected layer. A three-stream attention-aware network was proposed in \cite{chen2019three}, which extracts hierarchical features from RGB and depth inputs through two separate streams. Features are then progressively combined and selected via attention-aware blocks in the third stream. A new multi-scale multi-path fusion network with cross-modal interactions was proposed in \cite{chen2019multi}. Works \cite{liu2019salient} and \cite{huang2018rgbd} formulated a four-channel input by concatenating RGB and depth data. The input is later fed to a single-stream recurrent CNN and an FCN with short connections, respectively. The model in \cite{zhu2019pdnet} employed a subsidiary network to obtain depth features and used them to enhance the intermediate representation in an encoder-decoder architecture. Zhao \emph{et al.} \cite{zhao2019contrast} proposed a model that generates a contrast-enhanced depth map, which is later used as a prior map for feature enhancement in subsequent fluid pyramid integration. Fan \emph{et al.} \cite{fan2019rethinking} constructed a new RGB-D dataset called the Salient Person (SIP) dataset, and introduced a depth-depurator network to judge whether a depth map should be concatenated with the RGB image to formulate an input signal. Piao \emph{et al.} \cite{Piao2019depth} proposed a depth-induced multi-scale recurrent attention network, where the multi-scale fused features are re-weighted by a depth-induced vector and then processed by a recurrent attention module.
As concurrent works, Liu \emph{et al.} \cite{liu2020learning} proposed for RGB-D saliency detection a selective self-mutual attention mechanism inspired by the non-local model \cite{wang2018non}.
Zhang \emph{et al.} \cite{zhang2020select} designed a complimentary interaction module to discriminatively select representation from RGB and depth data, after which the learning was enhanced by a new compensation-aware loss.
Piao \emph{et al.} \cite{Piao2020a2dele} proposed attentive and adaptive depth distillation to learn an enhanced RGB salient object detector by transferring depth knowledge.
Zhang \emph{et al.} \cite{Zhang2020UCNet,zhang2020uncertainty} introduced the conditional variational autoencoder to model uncertainty in saliency annotation, which generates multiple potential saliency maps to be voted by a consensus module.

\begin{figure}
\includegraphics[width=0.485\textwidth]{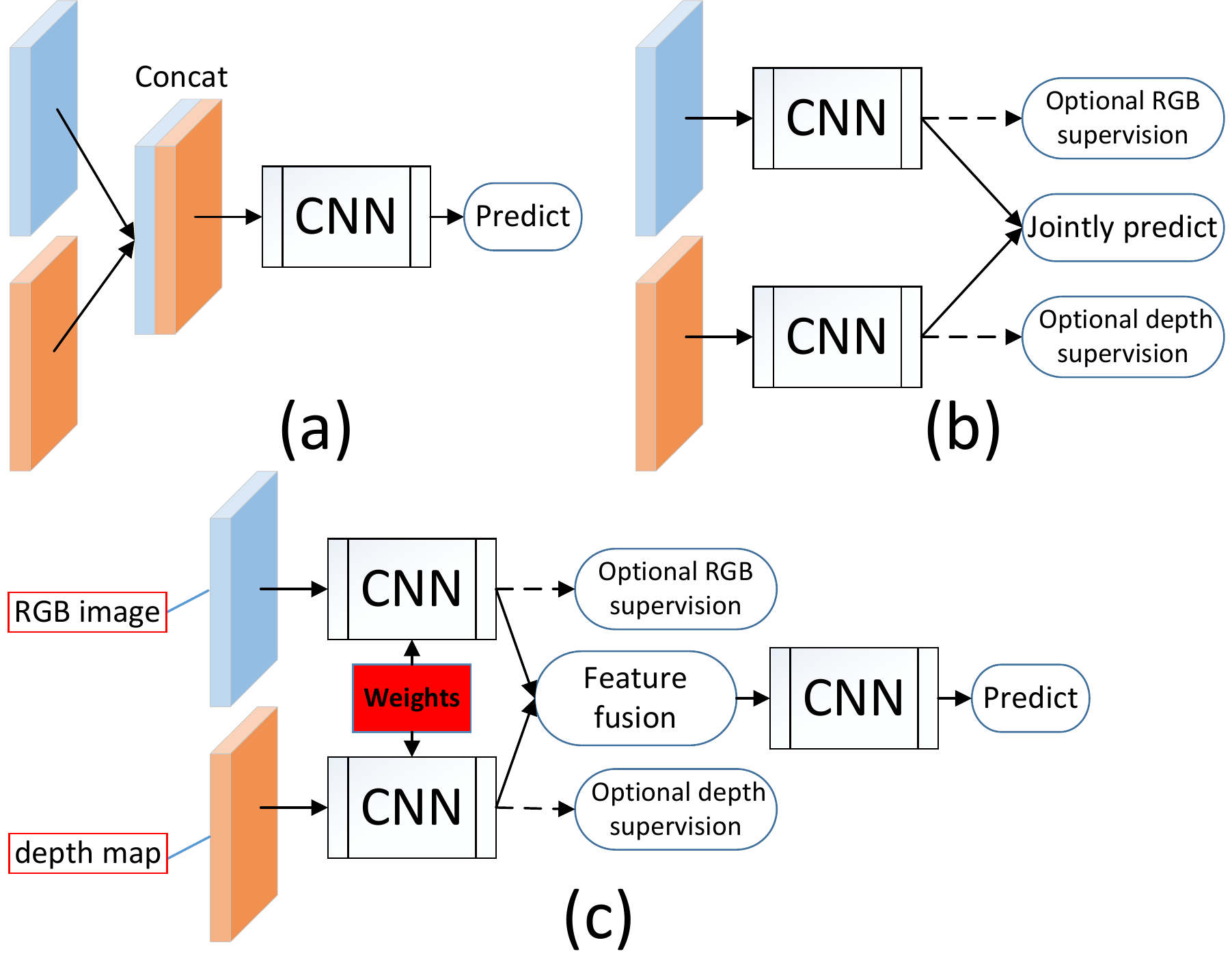}
\vspace{-15pt}
\caption{Typical schemes for RGB-D saliency detection. (a) Early-fusion. (b) Late-fusion. (c) Middle-fusion. }
\label{fig_sotascheme}
\end{figure}

\vspace{5pt}
\noindent\textbf{Categorization and discussion.}
Generally, as summarized by previous literature \cite{chen2018progressively,zhao2019contrast}, most of the above approaches can be divided into three categories: (a) early-fusion \cite{song2017depth,liu2019salient,huang2018rgbd,fan2019rethinking}, (b) late-fusion \cite{han2017cnns,wang2019adaptive} and (c) middle-fusion  \cite{chen2018progressively,zhu2019pdnet,chen2019three,chen2019multi,Piao2019depth,liu2020learning,zhang2020select}.
Fig. \ref{fig_sotascheme} (a)-(c) illustrate these three fusion strategies.
\emph{Early-fusion} (Fig. \ref{fig_sotascheme} (a)) uses simple concatenation to conduct input fusion. It may be difficult to capture the complementary interactions between the RGB and depth views, because these two types of information are blended in the very first stage but the supervision signal is finally far away from the blended input. The learning process is prone to local optima, where only either RGB or depth features are learned, and therefore may not guarantee improvement after view fusion. Besides, performing deep supervision for RGB and depth views individually is infeasible. This makes learning towards correct direction difficult.
\emph{Late-fusion} (Fig. \ref{fig_sotascheme} (b)) explicitly extracts RGB and depth features using two parallel networks. This ensures that both the RGB and depth views contribute to the final decision. Also, it is very straightforward to apply individual view-specific supervision in this scheme. However, the drawback is that this scheme fails to mine complex intrinsic correlations between the two views, \ie, the highly non-linear complementary rules.
\emph{Middle-fusion} (Fig. \ref{fig_sotascheme} (c)) complements (a) and (b), since both feature extraction and subsequent fusion are handled by relatively deep CNNs. As a consequence, high-level concepts can be learnt from both modalities and complex integration rules can be mined. Meanwhile, adding extra individual deep supervision for RGB and depth data is straightforward. 

The proposed \ourmodel~scheme falls under the middle-fusion category. However, unlike the aforementioned methods \cite{chen2018progressively,zhu2019pdnet,chen2019three,chen2019multi,Piao2019depth,liu2020learning,zhang2020select,fan2020bbs,zhai2020bifurcated}, where the two feature extraction streams are independent, we propose to utilize a Siamese architecture \cite{chopra2005learning}, where both the network architecture and weights are shared, as illustrated by the red part in Fig. \ref{fig_sotascheme} (c). This results in two major benefits: 1) Cross-modal knowledge-sharing becomes straightforward via joint learning; 2) Model parameters are largely reduced as only one shared network is needed, leading to a facilitated learning process.

\begin{figure*}
\includegraphics[width=\textwidth]{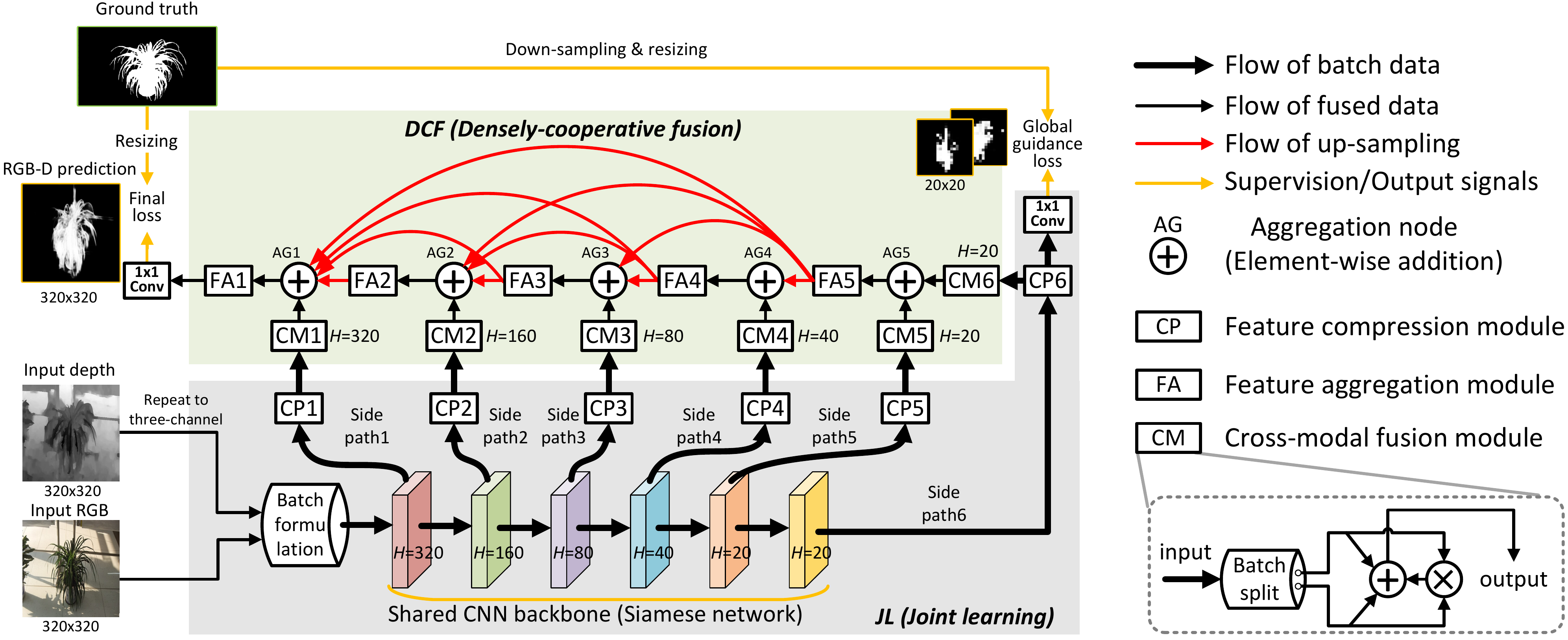}
\vspace{-15pt}
\caption{Block diagram of the proposed \ourmodel~framework for RGB-D SOD. The JL (joint learning) component is shown in gray, while the DCF (densely cooperative fusion) component is shown in light green. CP1$\sim$CP6: Feature compression modules. FA1$\sim$FA6: Feature aggregation modules. CM1$\sim$CM6: Cross-modal fusion modules. ``\emph{H}'' denotes the spatial size of output feature maps at a particular stage. See Section \ref{sec3} for details.}
\label{fig_blockdiagram}
\end{figure*}

\subsection{Siamese Networks in Computer Vision}\label{sec22}

The concept of ``Siamese network'' was introduced by Bromley \emph{et al.} \cite{bromley1994signature} in the 1990s for hand-written signature verification. In their work, two identical (\ie, ``Siamese'') neural networks with exactly the same parameters were introduced to handle two input signatures, and the feature vectors obtained were constrained by some distance measure during learning. This idea of a Siamese network was later extended to various computer vision tasks including face verification \cite{chopra2005learning,taigman2014deepface}, one-shot image recognition \cite{koch2015siamese}, stereo matching \cite{zagoruyko2015learning,zbontar2015computing,luo2016efficient,kendall2017end,khamis2018stereonet}, object tracking \cite{bertinetto2016fully,sun2019deep,guo2020siamcar,voigtlaender2020siam,chen2020siamese}, and semi-supervised video object segmentation \cite{cheng2018fast,wug2018fast,wang2019fast}. The essence of the Siamese network and why it can be applied lies in that it is suitable for learning general feature representations with a distance (or similarity) metric from two similar inputs, such as two face images \cite{chopra2005learning}, two image patches \cite{zagoruyko2015learning,zbontar2015computing}, a rectified pair of stereo images \cite{kendall2017end,khamis2018stereonet}, or a template image and a search image \cite{bertinetto2016fully}. After training, a Siamese network can be considered as an embedding serving in a comparison function. Recent works have attempted to manipulate features obtained from Siamese networks to formulate an elegant end-to-end framework \cite{kendall2017end,khamis2018stereonet}, or achieve more accurate feature learning and matching \cite{guo2020siamcar,voigtlaender2020siam,chen2020siamese}.
A comprehensive summary of the Siamese network is beyond the scope of this work. The readers can refer to the recently released survey work~\cite{marvasti2019deep} for more details.

Different from all the above works, in this paper, the Siamese network is aimed at exploiting saliency-aware cross-modal commonality and complementarity instead of matching or measuring distance. In other words, deep RGB and depth cues from the Siamese network are fused/merged, rather than compared, in order to achieve the desired RGB-D saliency prediction. It is worth noting that the Siamese network has not yet been introduced to multi-modal saliency detection, and even in the entire saliency detection community, there are only very few works utilizing Siamese networks.

\fdp{
\subsection{RGB-D Semantic Segmentation}\label{sec23}
RGB-D semantic segmentation is a close research area to RGB-D SOD. Although these two fields have different definitions on tasks, both of them aim at region segmentation. In contrast to segmenting salient object regions in RGB-D SOD, RGB-D semantic segmentation is to label all pixels within pre-defined categories given RGB-D inputs \cite{couprie2013indoor,Long2017Fully,gupta2014learning,hazirbas2016fusenet}. As a representative work, Shelhamer \emph{et al.} \cite{Long2017Fully} used FCNs to handle RGB-D semantic segmentation and experimented with early-fusion, \ie, by concatenating RGB and depth as a new input, as well as late-fusion, \ie, by averaging scores from RGB and HHA inputs \cite{gupta2014learning}. Existing RGB-D semantic segmentation techniques can be grouped into three types: 1) Treat depth as an additional input source and combine derived features with those from the RGB one \cite{couprie2013indoor,Long2017Fully,gupta2014learning,hazirbas2016fusenet,wang2016learning,li2016lstm,cheng2017locality,park2017rdfnet,deng2019rfbnet,chen2020bi}. 2) Recover 3D data from RGB-D sources and process with 3D/volumetric CNNs to handle appearances and geometries simultaneously \cite{qi20173d,song2017semantic}. 3) Utilize depth clues as auxiliary assistance to augment feature extraction from the RGB modality \cite{lin2017cascaded,wang2018depth,chen20193d,xing2020malleable,Chen2021spatial}, such as the depth-aware convolution/pooling \cite{wang2018depth}, 2.5D convolution \cite{xing2020malleable}, and S-Conv \cite{Chen2021spatial}. Also note that, among the models of the first type, further, they can be divided similarly (as in Fig. \ref{fig_sotascheme}) into three categories, \ie, early-fusion \cite{couprie2013indoor,Long2017Fully}, late-fusion \cite{Long2017Fully,gupta2014learning,li2016lstm,cheng2017locality}, and middle-fusion \cite{hazirbas2016fusenet,wang2016learning,park2017rdfnet,deng2019rfbnet,chen2020bi}, as summarized in recent literature \cite{park2017rdfnet,deng2019rfbnet}. This fact reveals a potential strong correlation between RGB-D SOD and semantic segmentation. We will further draw discussions between the proposed scheme and RGB-D semantic segmentation in Section \ref{sec47}.
}


\section{Methodology}\label{sec3}

The overall architecture of the proposed \ourmodel~is shown in Fig. \ref{fig_blockdiagram}. It follows the classic bottom-up/top-down strategy \cite{pinheiro2016learning}. For illustrative purpose, Fig. \ref{fig_blockdiagram} depicts an example backbone with six hierarchies that are common in the widely used VGG \cite{Simonyan14c} and ResNet \cite{He2015Deep}. The architecture consists of a JL component and a DCF component. The JL component conducts joint learning for the two modalities using a Siamese network. It aims to discover the commonality between these two views from a ``model-sharing'' perspective, since their information can be merged into the model parameters via back-propagation. As seen in Fig. \ref{fig_blockdiagram}, the hierarchical features jointly learned by the backbone are then fed to the subsequent DCF component. DCF is dedicated to feature fusion and its layers are constructed in a densely cooperative way. In this sense, the complementarity between RGB and depth modalities can be explored from a ``feature-integration'' perspective. To perform cross-view feature fusion, in the DCF component, we elaborately design a cross-modal fusion module (CM module in Fig. \ref{fig_blockdiagram}). Details about \ourmodel~will be given in the following sections.

\subsection{Joint Learning (JL)}\label{sec32}

As shown in Fig. \ref{fig_blockdiagram} (gray part), the inputs of the JL component are an RGB image together with its corresponding depth map. We first normalize the depth map into intervals [0, 255] and then convert it to a three-channel map through color mapping. In our implementation, we simply use the vanilla gray color mapping, which is equivalent to replicating the single channel map into three channels. Note that other color mappings \cite{al2016creating} or transformations, like the mean used in \cite{han2017cnns}, could also be considered for generating the three-channel representation. Next, the three-channel RGB image and transformed depth map are concatenated to formulate a \emph{batch}, so that the subsequent CNN backbone can perform parallel processing. Note that, unlike the early-fusion schemes previously mentioned, which often concatenate the RGB and depth inputs in the 3$^{rd}$ channel dimension, our scheme concatenates them in the 4$^{th}$ dimension, often called the batch dimension. For example, in our case, a transformed $320\times320\times3$ depth and a $320\times320\times3$ RGB map will formulate a batch of size $320\times320\times3\times2$, rather than $320\times320\times6$. 

The hierarchical features from the shared CNN backbone are then leveraged in a side-output way like \cite{hou2019deeply}. Since the side-output features have varied resolutions and channel numbers (usually the deeper, the more channels), we first employ a set of CP modules (CP1$\sim$CP6 in Fig. \ref{fig_blockdiagram}, practically implemented by convolutional layers plus ReLU non-linearities) to compress them to an identical, smaller number, denoted as $k$. We do this for the following two reasons: (1) Using a large number of feature channels for subsequent decoding is memory and computationally expensive and (2) unifying the number of feature channels facilitates various element-wise operations. Note that, here, the outputs from our CP modules are still batches, which are denoted as the thicker black arrows in Fig. \ref{fig_blockdiagram}.

Coarse localization can provide the basis for the following top-down refinement \cite{pinheiro2016learning}.
In addition, jointly learning the coarse localization guides the shared CNN to learn to extract independent hierarchical features
from the RGB and depth views simultaneously.
In order to enable the CNN backbone to coarsely locate the targets from both the RGB and depth views, we apply deep supervision to the JL component in the last hierarchy.
To achieve this, as shown in Fig. \ref{fig_blockdiagram}, we add a $(1 \times 1,1)$ convolutional layer after the CP6 module to achieve coarse prediction. The depth and RGB-associated outputs are supervised by the down-sampled ground truth map. The loss generated in this stage is called the global guidance loss $\mathcal{L}_{g}$.


\subsection{Densely Cooperative Fusion (DCF)}\label{sec33}

\begin{figure}
\includegraphics[width=0.485\textwidth]{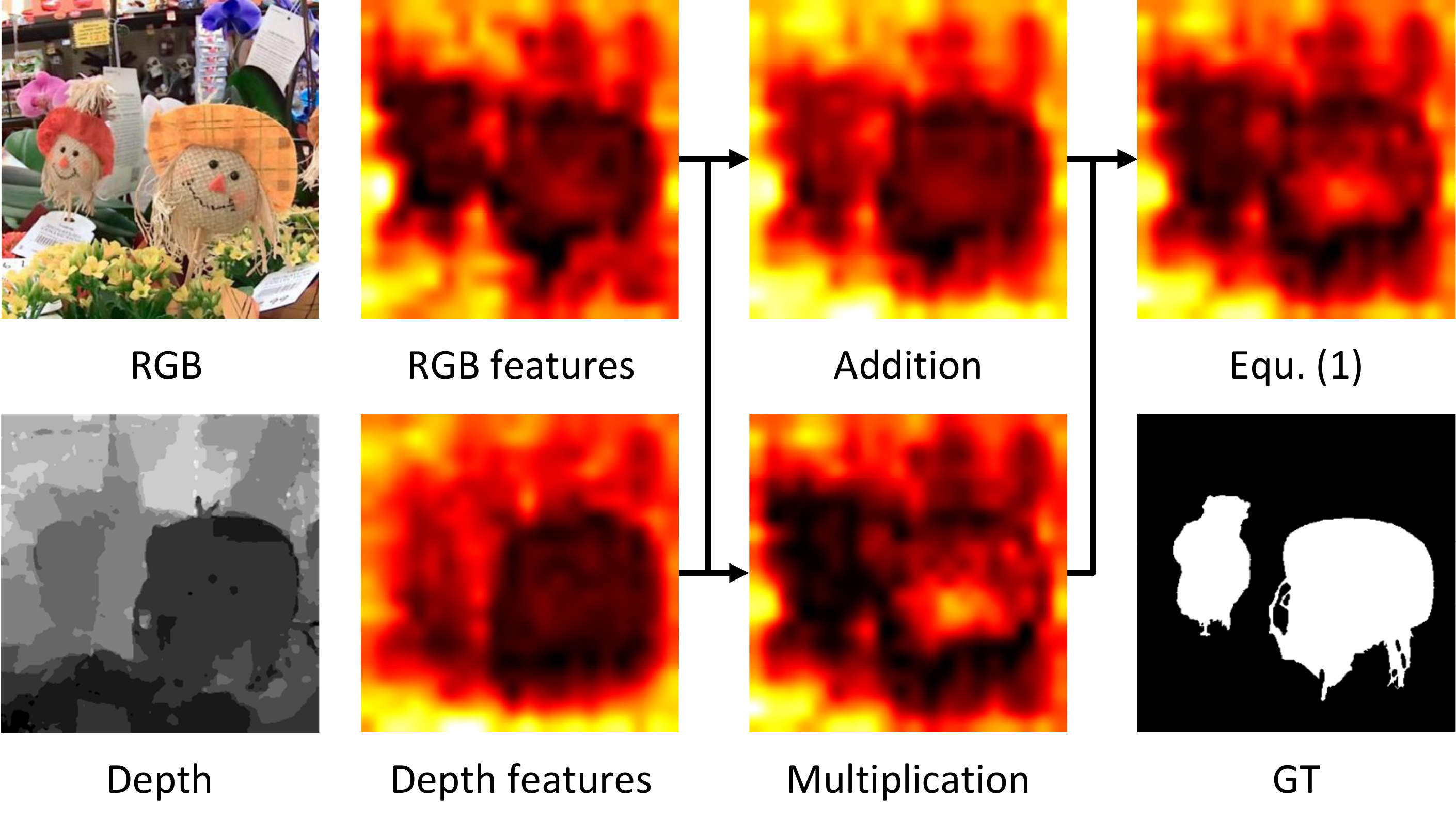}
\vspace{-15pt}
\caption{Intermediate feature visualization in CM6, where the RGB and depth features after batch split are visualized. Generally, addition and multiplication operations gather different cross-modal clues, making the features of both dolls enhanced after Equ. (\ref{equ_cm}).}
\label{fig_featurevisualize}
\end{figure}

As shown in Fig. \ref{fig_blockdiagram} (light green part), the output batch features from the CP modules contain depth and RGB information. They are fed to the DCF component, which can be considered a decoder that performs multi-scale cross-modal fusion. Firstly, we design a CM (cross-modal fusion) module to split and then merge the batch features (Fig. \ref{fig_blockdiagram}, bottom-right). This module first splits the batch data and then conducts ``addition and multiplication'' feature fusion, which we call \emph{cooperative fusion}. Mathematically, let a batch feature be denoted by $\{{X}_{rgb}, {X}_d\}$, where ${X}_{rgb}$, ${X}_d$ represent the RGB and depth feature tensors, each with $k$ channels, respectively. The CM module conducts the fusion as:

\begin{equation} \label{equ_cm}
CM(\{{X}_{rgb}, {X}_d\})={X}_{rgb} \oplus {X}_d \oplus ( {X}_{rgb} \otimes {X}_d),
\end{equation}

\noindent where ``$\oplus$'' and ``$\otimes$'' denote element-wise addition and multiplication. The blended features output from the CM modules are still made up of $k$ channels. Equ. (\ref{equ_cm}) enforces explicit feature fusion indicated by ``$\oplus$'' and ``$\otimes$'', where ``$\oplus$'' exploits \emph{feature complementarity} and ``$\otimes$'' puts more emphasis on \emph{feature commonality}. These two properties intuitively gather different clues, as
shown in Fig. \ref{fig_featurevisualize}, and are generally important in cross-view fusion. On the other hand, Equ. (\ref{equ_cm}) can also be deemed as a kind of mutual residual attention \cite{wang2017residual} combining ``$A+A\otimes B$'' and ``$B+B\otimes A$'', where $A$ and $B$ are the two types of features (\ie, ${X}_{rgb}, {X}_d$) each of which attends the elements in the other in a residual way.

One may argue that the above CM module could be replaced by channel concatenation, which generates $2k$-channel concatenated features. However, we find such a choice tends to result in the learning process being trapped in a local optimum, where it becomes biased towards only RGB information. The reason seems to be that the channel concatenation does indeed involve feature selection rather than explicit feature fusion. This leads to degraded learning outcomes, where RGB features easily dominate the final prediction. Note that, as will be shown in Section \ref{sec44}, solely using RGB input can also achieve fairly good performance in the proposed framework. Comparisons between our CM modules and concatenation will be given in Section \ref{sec44}. 

\begin{figure}
\includegraphics[width=0.48\textwidth]{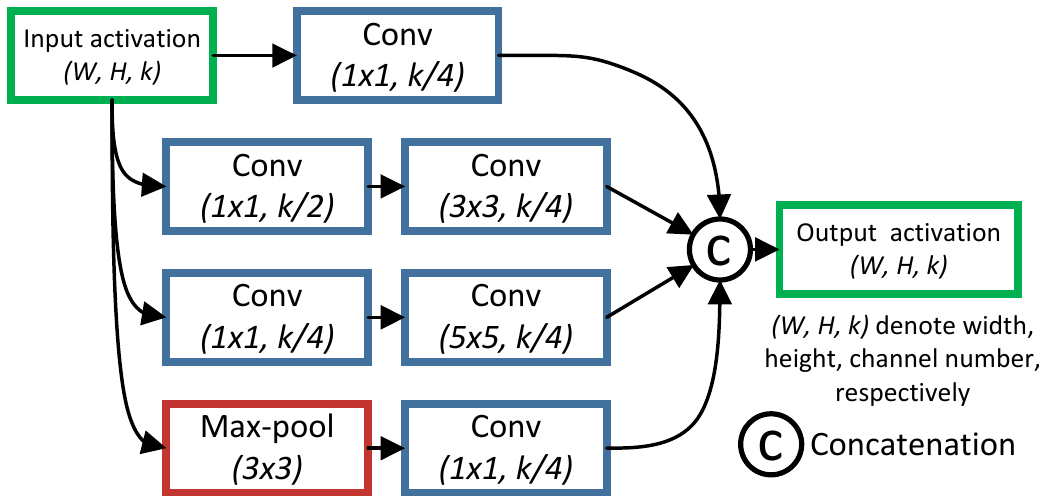}
\vspace{-10pt}
\caption{Inception structure used for the FA modules in Fig. \ref{fig_blockdiagram}. All convolutional and max-pooling layers have stride 1, therefore maintaining spatial feature sizes. Unlike the original Inception module \cite{szegedy2015going}, we adapt it to have the same input/output channel number $k$.}
\label{fig_inception}
\end{figure}

As shown in Fig. \ref{fig_blockdiagram}, the fused features from CM1$\sim$CM6 are fed to a decoder augmented with a dense connection \cite{huang2017densely}. Using the dense connection promotes the blending of depth and RGB features at various scales. Therefore, unlike the traditional UNet-like decoder \cite{ronneberger2015u}, an aggregation module FA takes inputs from all levels deeper than itself. Specifically, FA denotes a feature aggregation module performing non-linear aggregation and transformation. To this end, we use the Inception module \cite{szegedy2015going} shown in Fig. \ref{fig_inception}, which performs multi-level convolutions with filter size $1\times1,3\times3,5\times5$, and max-pooling. Note that the FA module in our framework is quite flexible. Other modules (\eg, \cite{hu2018squeeze,wang2018non,woo2018cbam,gao2019res2net}) may also be considered in the future to improve the performance.

Finally, the FA module with the finest features is denoted as FA1, whose output is then fed to a $(1 \times 1,1)$ convolutional layer to generate the final activation and then  ultimately the saliency map. This final prediction is supervised by the resized ground truth (GT) map during training. We denote the loss generated in this stage as $\mathcal{L}_{f}$.

\subsection{Loss Function}\label{sec34}

The total loss function of our scheme is composed of the global guidance loss $\mathcal{L}_{g}$ and final loss $\mathcal{L}_{f}$. Assume that $G$ denotes supervision from the ground truth, $S^c_{rgb}$ and $S^{c}_{d}$ denote the coarse prediction maps contained in the batch after module CP6, and $S^{f}$ is the final prediction after module FA1. The total loss function is defined as:

\begin{equation} \label{equ_loss}
\mathcal{L}_{total}=\mathcal{L}_{f}(S^{f}, G) + \lambda \sum_{x \in \{rgb, d\}}\mathcal{L}_{g}(S^c_{x}, G),
\end{equation}

\noindent where $\lambda$ balances the emphasis of global guidance, and we adopt the widely used cross-entropy loss for $\mathcal{L}_{g}$ and $\mathcal{L}_{f}$ as:

\begin{equation} \label{equ_cel}
\mathcal{L}(S,G)=-\sum_{i}[G_i \log (S_i) + (1-G_i) \log (1-S_i)],
\end{equation}

\noindent where $i$ denotes pixel index, and $S \in \{S^c_{rgb}, S^c_{d}, S^{f}\}$.

\subsection{Bridging between RGB and RGB-D SOD}\label{sec35}

Since the RGB and depth modalities share the same master CNN backbone for feature extraction in \ourmodel, it is easy to adapt \ourmodel~to a single modality (\eg,~RGB or depth) by replacing all the batch-related operations, such as the batch formulation and CM modules in Fig. \ref{fig_blockdiagram}, with identity mappings while keeping all the other settings unchanged, including the dense decoder and deep supervision. In this way, one can get a full-resolution saliency estimation result from either RGB or depth input. As a consequence, we can bridge between RGB and RGB-D SOD in terms of a data perspective in the training phase of \ourmodel. The underlying motivation is to use more RGB data to augment the generalizability of the JL component in \ourmodel, as the JL component is shared by both the RGB and depth views. The newly incorporated RGB-based knowledge could help improve the Siamese network regarding both RGB and depth modalities.

\begin{figure}\centering
\includegraphics[width=0.5\textwidth]{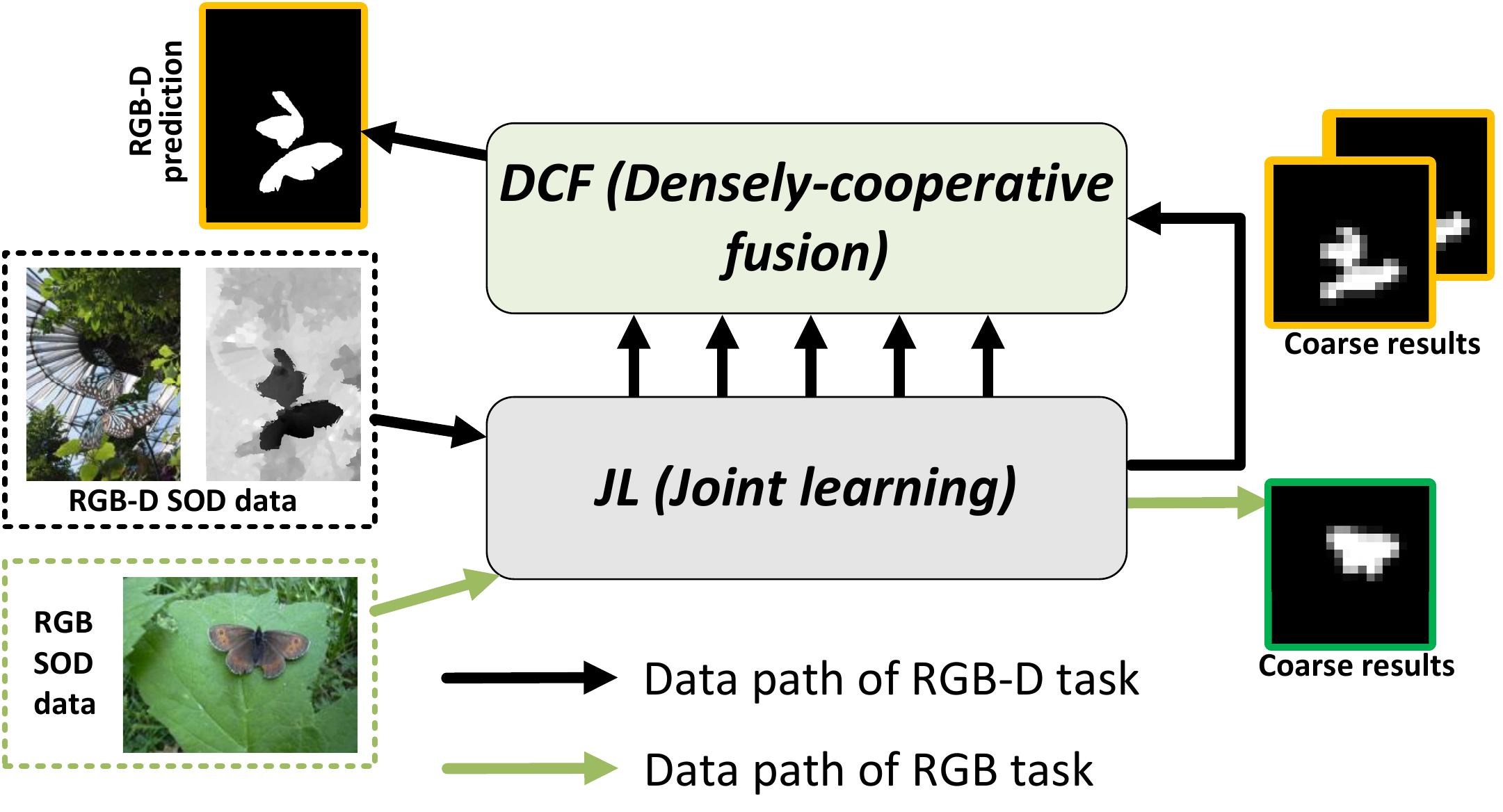}
\vspace{-15pt}
\caption{Bridging the RGB and RGB-D SOD tasks through \ourmodel, where the JL and DCF components are detailed in Fig. \ref{fig_blockdiagram}. During training, the network of \ourmodel~is simultaneously trained/optimized in an online manner for both tasks.}
\label{fig_bridge}
\end{figure}

To this end, we propose to further extend the JL component in a multi-task manner by considering RGB and RGB-D SOD as two simultaneous tasks. As shown in Fig. \ref{fig_bridge}, the JL component is shared across RGB and RGB-D SOD, and is jointly optimized by the data sources (\ie, training datasets) of these two tasks. Note that the RGB SOD datasets that can currently be used for training are much larger than the RGB-D ones, leading to a potential boost in generalizability. Practically, we obtain a coarse saliency map for the RGB SOD task from the JL component, and therefore the overall loss function, $\mathcal{L}_{total}^*$ in this case, can be written as the sum of the losses for the two tasks:

\begin{equation} \label{equ_joint_loss}
\mathcal{L}_{total}^*=\mathcal{L}_{f}(S^{f}, G) + \lambda \sum_{x \in \{ rgb, d, rgb*\}}\mathcal{L}_{g}(S^c_{x}, G),
\end{equation}

\noindent where $S^c_{rgb*}$ denotes the obtained coarse saliency map corresponding to the RGB SOD task, while other notations are defined the same as in Equ. (\ref{equ_loss}). More specifically, an RGB image for the RGB SOD task is concatenated with the RGB-D data in the batch dimension to formulate a single batch, which is then fed to the CNN backbone. The coarse prediction associated with the RGB SOD task is obtained by batch splitting and then supervised by the corresponding ground truth. Following the same supervision of $S_{rgb}^c$ in the RGB-D task, we use the standard cross-entropy loss for the RGB SOD task. Finally, it is worth noting that our above scheme aims at leveraging additional RGB SOD data to augment RGB-D SOD, while in contrast the recent work \cite{li2020is} aims at using additional RGB-D SOD data for training in order to augment RGB SOD.

\begin{table*}[t!]
    \renewcommand{\arraystretch}{0.8}
    \caption{Quantitative measures: S-measure ($S_\alpha$) \cite{Fan2017}, max F-measure ($F_{\beta}^{\textrm{max}}$) \cite{Borji2015TIP}, max E-measure ($E_{\phi}^{\textrm{max}}$) \cite{fan2018enhanced} and MAE ($M$) \cite{Perazzi2012} of SOTA methods and the proposed \ourmodel~and $\ourmodel^*$ (jointly trained with both RGB-D and RGB datasets) on six RGB-D SOD datasets.
    The best and second best results are highlighted in \textbf{bold} and \emph{italics}, respectively$^{\rm a}$.}\label{tab_sota}
   \centering
    \footnotesize
    \setlength{\tabcolsep}{1mm}
    \begin{tabular}{p{0.8mm}p{0.8mm}r||c|c|c|c|c|c|c|c|c|c|c|c|c|c||c|c}
    \hline
    &  & & \multicolumn{5}{c|}{Traditional}
    &\multicolumn{11}{c}{Deep Learning} \\
    \cline{4-19}
            && Metric  & \tabincell{c}{ACSD\\\cite{ju2014depth}} & \tabincell{c}{LBE\\\cite{feng2016local}} & \tabincell{c}{DCMC\\\cite{cong2016saliency}}  & \tabincell{c}{MDSF\\\cite{song2017depth}} & \tabincell{c}{SE\\\cite{guo2016salient}} & \tabincell{c}{DF\\\cite{qu2017rgbd}} & \tabincell{c}{AFNet\\\cite{wang2019adaptive}} & \tabincell{c}{CTMF\\\cite{han2017cnns}} & \tabincell{c}{MMCI\\\cite{chen2019multi}} & \tabincell{c}{PCF\\\cite{chen2018progressively}} & \tabincell{c}{TANet\\\cite{chen2019three}} & \tabincell{c}{CPFP\\\cite{zhao2019contrast}} & \tabincell{c}{DMRA\\\cite{Piao2019depth}}& \tabincell{c}{D3Net\\\cite{fan2019rethinking}}& \tabincell{c}{\ourmodel\\Ours}&\tabincell{c}{\ourmodel$^*$\\Ours}\\
    \hline
    \hline
          \multirow{4}{*}{\begin{sideways}\textit{NJU2K}\end{sideways}} & \multirow{4}{*}{\begin{sideways}\cite{ju2014depth}\end{sideways}} & $S_\alpha\uparrow$       &0.699&0.695&0.686&0.748&0.664&0.763&0.772&0.849&0.858&0.877&0.878&0.879&0.886&0.895&\emph{0.903}&\textbf{0.911}\\
                                                                        && $F_{\beta}^{\textrm{max}}\uparrow$       &0.711&0.748&0.715&0.775&0.748&0.804&0.775&0.845&0.852&0.872&0.874&0.877&0.886&0.889&\emph{0.903}&\textbf{0.913}\\
                                                                        && $E_{\phi}^{\textrm{max}}\uparrow$       &0.803&0.803&0.799&0.838&0.813&0.864&0.853&0.913&0.915&0.924&0.925&0.926&0.927&0.932&\emph{0.944}&\textbf{0.948}\\
                                                                        && $M\downarrow$       &0.202&0.153&0.172&0.157&0.169&0.141&0.100&0.085&0.079&0.059&0.060&0.053&0.051&0.051&\emph{0.043}&\textbf{0.040}\\
    \hline
        \multirow{4}{*}{\begin{sideways}\textit{NLPR}\end{sideways}} & \multirow{4}{*}{\begin{sideways}\cite{peng2014rgbd}\end{sideways}}& $S_\alpha\uparrow$
        &0.673&0.762&0.724&0.805&0.756&0.802&0.799&0.860&0.856&0.874&0.886&0.888&0.899&0.906&\emph{0.925}&\textbf{0.926}\\
                                                                        && $F_{\beta}^{\textrm{max}}\uparrow$      &0.607&0.745&0.648&0.793&0.713&0.778&0.771&0.825&0.815&0.841&0.863&0.867&0.879&0.885&\emph{0.916}&\textbf{0.917}\\
                                                                        && $E_{\phi}^{\textrm{max}}\uparrow$     &0.780&0.855&0.793&0.885&0.847&0.880&0.879&0.929&0.913&0.925&0.941&0.932&0.947&0.946&\emph{0.962}&\textbf{0.964}\\
                                                                        && $M\downarrow$       &0.179&0.081&0.117&0.095&0.091&0.085&0.058&0.056&0.059&0.044&0.041&0.036&0.031&0.034&\textbf{0.022}&\emph{0.023}\\

    \hline
        \multirow{4}{*}{\begin{sideways}\textit{STERE}\end{sideways}}& \multirow{4}{*}{\begin{sideways}\cite{niu2012leveraging}\end{sideways}} & $S_\alpha\uparrow$ &0.692&0.660&0.731&0.728&0.708&0.757&0.825&0.848&0.873&0.875&0.871&0.879&0.886&0.891&\emph{0.905}&\textbf{0.911}\\
                                                                        && $F_{\beta}^{\textrm{max}}\uparrow$       &0.669&0.633&0.740&0.719&0.755&0.757&0.823&0.831&0.863&0.860&0.861&0.874&0.886&0.881&\emph{0.901}&\textbf{0.907}\\
                                                                        && $E_{\phi}^{\textrm{max}}\uparrow$     &0.806&0.787&0.819&0.809&0.846&0.847&0.887&0.912&0.927&0.925&0.923&0.925&0.938&0.930&\emph{0.946}&\textbf{0.949}\\
                                                                        && $M\downarrow$       &0.200&0.250&0.148&0.176&0.143&0.141&0.075&0.086&0.068&0.064&0.060&0.051&0.047&0.054&\emph{0.042}&\textbf{0.039}\\
    \hline
        \multirow{4}{*}{\begin{sideways}\textit{RGBD135}\end{sideways}} & \multirow{4}{*}{\begin{sideways}\cite{cheng2014depth}\end{sideways}}  & $S_\alpha\uparrow$       &0.728&0.703&0.707&0.741&0.741&0.752&0.770&0.863&0.848&0.842&0.858&0.872&0.900&0.904&\emph{0.929}&\textbf{0.936}\\
                                                                        && $F_{\beta}^{\textrm{max}}\uparrow$     &0.756&0.788&0.666&0.746&0.741&0.766&0.728&0.844&0.822&0.804&0.827&0.846&0.888&0.885&\emph{0.919}&\textbf{0.929}\\
                                                                        && $E_{\phi}^{\textrm{max}}\uparrow$       &0.850&0.890&0.773&0.851&0.856&0.870&0.881&0.932&0.928&0.893&0.910&0.923&0.943&0.946&\emph{0.968}&\textbf{0.975}\\
                                                                        && $M\downarrow$       &0.169&0.208&0.111&0.122&0.090&0.093&0.068&0.055&0.065&0.049&0.046&0.038&0.030&0.030&\emph{0.022}&\textbf{0.021}\\
                                                                                                                                                                                                      \hline
        \multirow{4}{*}{\begin{sideways}\textit{LFSD}\end{sideways}} & \multirow{4}{*}{\begin{sideways}\cite{li2014saliency}\end{sideways}}  & $S_\alpha\uparrow$       &0.734&0.736&0.753&0.700&0.698&0.791&0.738&0.796&0.787&0.794&0.801&0.828&0.847&0.832&\emph{0.862}&\textbf{0.863}\\
                                                                        && $F_{\beta}^{\textrm{max}}\uparrow$      &0.767&0.726&0.817&0.783&0.791&0.817&0.744&0.792&0.771&0.779&0.796&0.826&0.857&0.819&\textbf{0.866}&\emph{0.862}\\
                                                                        && $E_{\phi}^{\textrm{max}}\uparrow$       &0.837&0.804&0.856&0.826&0.840&0.865&0.815&0.865&0.839&0.835&0.847&0.872&0.901&0.864&\textbf{0.901}&\emph{0.900}\\
                                                                        && $M\downarrow$       &0.188&0.208&0.155&0.190&0.167&0.138&0.134&0.119&0.132&0.112&0.111&0.088&0.075&0.099&\emph{0.071}&\textbf{0.071}\\
     \hline
        \multirow{4}{*}{\begin{sideways}\textit{SIP}\end{sideways}} & \multirow{4}{*}{\begin{sideways}\cite{fan2019rethinking}\end{sideways}} & $S_\alpha\uparrow$       &0.732&0.727&0.683&0.717&0.628&0.653&0.720&0.716&0.833&0.842&0.835&0.850&0.806&0.864&\emph{0.879}&\textbf{0.892}\\
                                                                        && $F_{\beta}^{\textrm{max}}\uparrow$       &0.763&0.751&0.618&0.698&0.661&0.657&0.712&0.694&0.818&0.838&0.830&0.851&0.821&0.862&\emph{0.885}&\textbf{0.900}\\
                                                                        && $E_{\phi}^{\textrm{max}}\uparrow$       &0.838&0.853&0.743&0.798&0.771&0.759&0.819&0.829&0.897&0.901&0.895&0.903&0.875&0.910&\emph{0.923}&\textbf{0.949}\\                                                                     && $M\downarrow$       &0.172&0.200&0.186&0.167&0.164&0.185&0.118&0.139&0.086&0.071&0.075&0.064&0.085&0.063&\emph{0.051}&\textbf{0.046}\\
\hline
\end{tabular}
\begin{flushleft}
\justifying
\fkr{$^{\rm a}$We also have implemented Pytorch versions of \ourmodel~with different backbones, \ie, ResNet-101, ResNet-50, and VGG-16. They all achieve SOTA performance comparing with the models in this table. Generally, due to differences between deep learning platforms, the Pytorch versions are found to perform moderately better than the Caffe implementation. Results and models can be found on our project site.}
\end{flushleft}\vspace{-0.5cm}
\end{table*}
\section{Experiments}\label{sec4}

\subsection{Datasets and Metrics}\label{sec41}

Experiments are conducted on six classic RGB-D benchmark datasets: NJU2K \cite{ju2014depth} (2,000 samples), NLPR \cite{peng2014rgbd} (1,000 samples), STERE \cite{niu2012leveraging} (1,000 samples), RGBD135 \cite{cheng2014depth} (135 samples), LFSD \cite{li2014saliency} (100 samples), and SIP \cite{fan2019rethinking} (929 samples), as well as the recently proposed dataset DUT-RGBD \cite{Piao2019depth} (only testing subset, 400 samples). Following \cite{zhao2019contrast}, we choose the same 700 samples from NLPR and 1,500 samples from NJU2K, resulting in 2,200 samples in total, to train our algorithms. The remaining samples are used for testing. Besides, when jointly training \ourmodel~with both RGB-D and RGB sources, for the RGB dataset we use the training set (10,553 images) of DUTS \cite{wang2017learning}, which is currently the largest saliency detection benchmark with an explicit training/test evaluation protocol, commonly used for training recent RGB SOD models \cite{zhang2018progressive,feng2019attentive,wang2019salient}. For fair comparison, we apply the model trained on this training set to other datasets.

For evaluation purposes, we adopt five widely used metrics, namely precision-recall curve \cite{Achanta2009,cheng2015global,Borji2015TIP}, S-measure ($S_\alpha$) \cite{Fan2017}, maximum F-measure ($F_{\beta}^{\textrm{max}}$) \cite{Borji2015TIP,hou2019deeply}, maximum E-measure ($E_\phi^{\textrm{max}}$) \cite{fan2018enhanced}, and MAE ($M$) \cite{Perazzi2012,Borji2015TIP}. Given a saliency map $S_{map}$ and the ground truth map $G$, the definitions for these metrics are as follows:

\begin{enumerate}
\item \emph{Precision-Recall (PR)} \cite{Achanta2009,cheng2015global,Borji2015TIP} is defined as:
\begin{equation} \label{equ9}
\begin{array}{l}
\textrm{Precision}(T)=\frac{|M(T)\cap{G}|}{|M(T)|},~~\textrm{Recall}(T) =\frac{|M(T)\cap{G}|}{|G|},
\end{array}
\end{equation}
where $M(T)$ is the binary mask obtained by directly thresholding the saliency map $S_{map}$ with the threshold $T$, and $|\cdot|$ is the total area of the mask(s) inside the map. By varying $T$, a precision-recall curve can be obtained.

\item \emph{S-measure ($S_{\alpha}$)}~\cite{Fan2017} was proposed to measure the spatial structure similarities of saliency
maps:
\begin{equation} \label{equ13}
S_{\alpha}=\alpha \ast S_{o} + (1- \alpha)\ast S_{r},
\end{equation}
where $\alpha$ is a balance parameter between object-aware structural similarity $S_{o}$ and region-aware structural similarity $S_{r}$. We set $\alpha=0.5$, as in \cite{Fan2017}.

\item \emph{F-measure} ($F_{\beta}$) \cite{Borji2015TIP,hou2019deeply} is defined as:
\begin{equation} \label{equ10}
F_{\beta}=\frac{(1+\beta^{2})\textrm{Precision} \cdot \textrm{Recall}}{\beta^{2}\cdot {\textrm{Precision}}+\textrm{Recall}},
\end{equation}
where $\beta $ is the weight between the precision and the recall. $\beta^{2}=0.3$ is usually set since the precision is often weighted more than the recall. In order to get a single-valued score, a threshold is often applied to binarize a saliency map into a foreground mask map. In this paper, we report the maximum F-measure, \ie,~$F_{\beta}^{\textrm{max}}$, computed from the precision-recall curve by running all threshold values (\ie,~[0, 255]).

\item \emph{E-measure ($E_{\phi}$)} was proposed in \cite{fan2018enhanced} as an enhanced-measure for comparing two binary maps. This metric first aligns two binary maps according to their global means and then computes the local pixel-wise correlation. Finally, an enhanced alignment matrix $\phi$ of the same size as the binary maps is obtained and $E_{\phi}$ is defined as:

\begin{equation} \label{equ10_2}
E_{\phi}=\frac{1}{W \cdot H}\sum_{x=1}^{W}\sum_{y=1}^{H}\phi(x,y),
\end{equation}
where $\phi(x,y)$ denotes the matrix entry at pixel location $(x,y)$. $W$ and $H$ are the width and height of $S_{map}$. The range of $E_{\phi}$ lies in intervals [0, 1]. To extend it for comparing a non-binary saliency map against a binary ground truth map, we follow a similar strategy to $F_{\beta}^{\textrm{max}}$. Specifically, we first binarize a saliency map into a series of foreground maps using all possible threshold values in [0, 255], and report the maximum E-measure, \ie,~$E_{\phi}^{\textrm{max}}$, among them.

\begin{figure*}[t!]
\centering
\includegraphics[width=.98\textwidth]{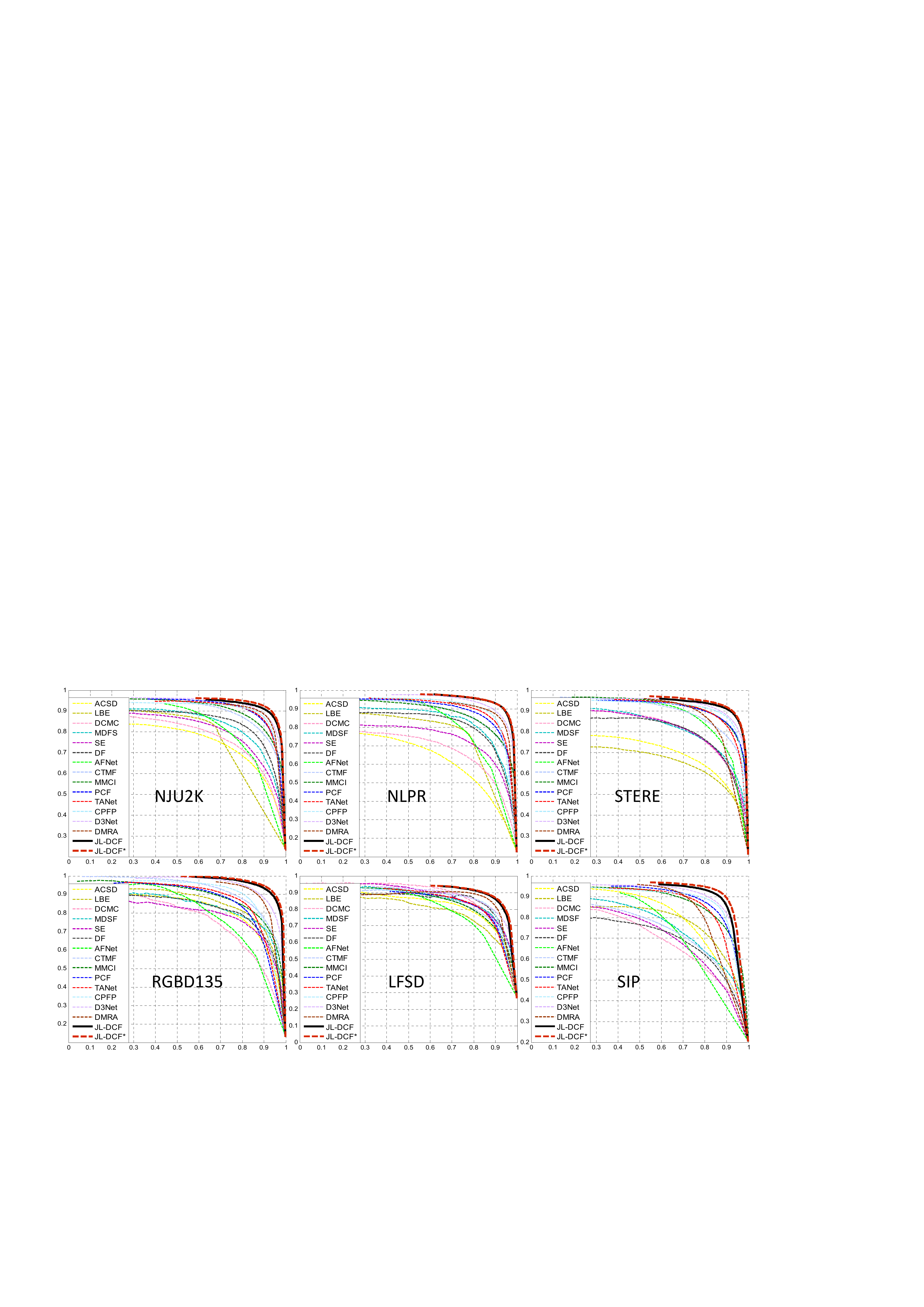}
\vspace{-10pt}
\caption{Precision-recall curves of SOTA methods and the proposed \ourmodel~and \ourmodel$^*$ across six datasets.}
\label{fig_sotapr}
\end{figure*}

\item \emph{Mean Absolute Error (MAE)} \cite{Perazzi2012,Borji2015TIP} is defined as:
\begin{equation} \label{equ11}
M=\frac{1}{W \cdot H}\sum_{x=1}^{W}\sum_{y=1}^{H}|S_{map}(x,y)-G(x,y)|,
\end{equation}
where $S_{map}(x,y)$ and $G(x,y)$ correspond to the saliency value and ground truth value at pixel location $(x,y)$. $W$ and $H$ are the width and height of the saliency map $S_{map}$.

\end{enumerate}

\noindent In summary, for the five metrics above, higher precision-recall curves, $S_{\alpha}$, $F_{\beta}^{\textrm{max}}$, $E_{\phi}^{\textrm{max}}$, and lower $M$ indicate better performance.

\begin{table}\centering
\caption{Details of the  two extra convolutional (Conv.) layers inserted into the \emph{side} \emph{path1}$\sim$\emph{path6} (for both the VGG-16 and ResNet-101 configuration). Parameters in the below brackets from left to right are: kernel size, channel number, stride, dilation rate, and padding.}\label{tab_sideconv}
{
\renewcommand{\tabcolsep}{5.5mm}
\begin{tabular}{c||c||c}
\hline
\emph{No.}\textbackslash Layers     & 1$^{st}$ Conv. layer & 2$^{nd}$ Conv. layer \\
\hline
\emph{Side path1}                          & (3, 128, 1, 1, 1) & (3, 128, 1, 1, 1)\\
\hline
\emph{Side path2}                          & (3, 128, 1, 1, 1) & (3, 128, 1, 1, 1)\\
\hline
\emph{Side path3}                          & (5, 256, 1, 1, 2) & (5, 256, 1, 1, 2)\\
\hline
\emph{Side path4}                          & (5, 256, 1, 1, 2) & (5, 256, 1, 1, 2)\\
\hline
\emph{Side path5}                          & (5, 512, 1, 1, 2) & (5, 512, 1, 1, 2)\\
\hline
\emph{Side path6}                          & (7, 512, 1, 2, 6) & (7, 512, 1, 2, 6)\\
\hline
\end{tabular}
}
\end{table}


\subsection{Implementation Details}\label{sec42}

The proposed \ourmodel~scheme is generally independent from the network backbone. In this work, we implement two versions of \ourmodel~based on VGG-16 \cite{Simonyan14c} and ResNet-101 \cite{He2015Deep}, respectively. We fix the input size of the network as $320 \times 320 \times 3$. Simple gray color mapping is adopted to convert a depth map into a three-channel map.

\textbf{VGG-16 configuration:} For the VGG-16 with fully connected layers removed and meanwhile having 13 convolutional layers, the \emph{side} \emph{path1}$\sim$\emph{path6} are successively connected to \emph{conv1\_2}, \emph{conv2\_2}, \emph{conv3\_3}, \emph{conv4\_3}, \emph{conv5\_3}, and \emph{pool5}. Inspired by \cite{hou2019deeply}, we add two extra convolutional layers into \emph{side} \emph{path1}$\sim$\emph{path6}.
To augment the resolution of the coarsest feature maps from \emph{side} \emph{path6}, while at the same time preserving the receptive field, we let \emph{pool5} have a stride of 1 and instead use dilated convolution \cite{chen2017deeplab} with a rate of 2 for the two extra side convolutional layers. Details of the extra side convolutional layers are given in Table \ref{tab_sideconv}.
Generally, the coarsest features produced by our modified VGG-16 backbone have a spatial size of $20 \times 20$, as in Fig. \ref{fig_blockdiagram}.

\textbf{ResNet-101 configuration:} Similar to the VGG-16 case above, the spatial size of the coarsest features produced by our modified ResNet backbone is also $20 \times 20$. As the first convolutional layer of ResNet already has a stride of 2, the features from the shallowest level have a spatial size of $160 \times 160$. To obtain the full size ($320\times320$) features without trivial up-sampling, we borrow the \emph{conv1\_1} and \emph{conv1\_2} layers from VGG-16 for feature extraction. \emph{Side} \emph{path1}$\sim$\emph{path6} are connected to \emph{conv1\_2}, and \emph{conv1}, \emph{res2c}, \emph{res3b3}, \emph{res4b22}, \emph{res5c} of the ResNet, respectively. We also change the stride of the \emph{res5a} block from 2 to 1, but subsequently use dilated convolution with rate 2.

\textbf{Decoder configuration:} All CP modules in Fig. \ref{fig_blockdiagram} are $3 \times 3$ convolutions with $k=64$ filters, and all FA modules are Inception modules. Up-sampling is achieved by simple bilinear interpolation. As depicted in Fig. \ref{fig_blockdiagram}, to align the feature sizes in the decoder, the output from an FA module is up-sampled by various factors. In an extreme case, the output from FA5 is up-sampled by a factor of $2$, $4$, $8$, and $16$. The final output from FA1 has a spatial size of $320 \times 320$, which is identical to the initial input.

\begin{figure*}[t!]
\includegraphics[width=.98\textwidth]{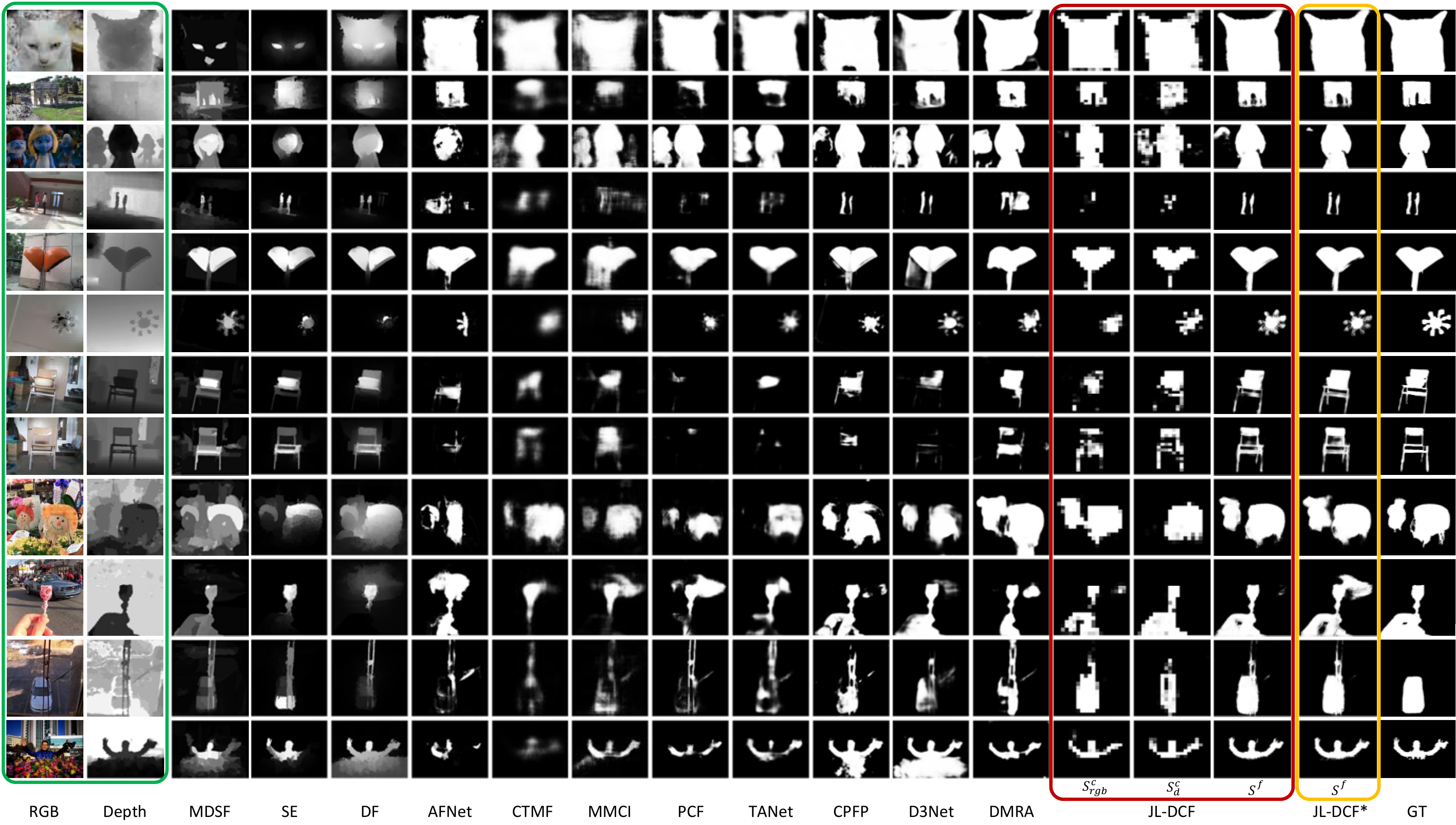}
\vspace{-10pt}
\caption{Visual comparisons of \ourmodel~(trained with only RGB-D data) \fdp{and \ourmodel$^*$ (trained with both RGB-D and RGB data)} with SOTA RGB-D saliency models. The jointly learned coarse prediction maps ($S^c_{rgb}$ and $S^c_{d}$) from RGB and depth are also shown together with the final maps ($S^{f}$) of \ourmodel.}
\label{fig_sotavisual}
\end{figure*}

\textbf{Training setup:} We implement \ourmodel~on Caffe \cite{jia2014caffe}. During training, the backbone~\cite{Simonyan14c,He2015Deep} is initialized by the pre-trained model, 
and other layers are randomly initialized. We fine-tune the entire network through end-to-end joint learning. Training data is augmented by mirror reflection to generate double the amount of data. The momentum parameter is set as 0.99, the learning rate is set to $lr=10^{-9}$, and the weight decay is 0.0005. The weight $\lambda$ in Eq. (\ref{equ_loss}) is set as 256 (=16$^2$) to balance the loss between the low- and high-resolution predictions. Stochastic Gradient Descent learning is adopted and accelerated by an NVIDIA 1080Ti GPU.
The training time is about 20 hours/18 hours for 40 epochs under the ResNet-101/VGG-16 configuration. Incorporating RGB data for multi-task training of the same epochs on ResNet-101 requires seven more hours.

\subsection{Comparisons to SOTAs}\label{sec43}

\begin{figure}\centering
\includegraphics[width=0.38\textwidth]{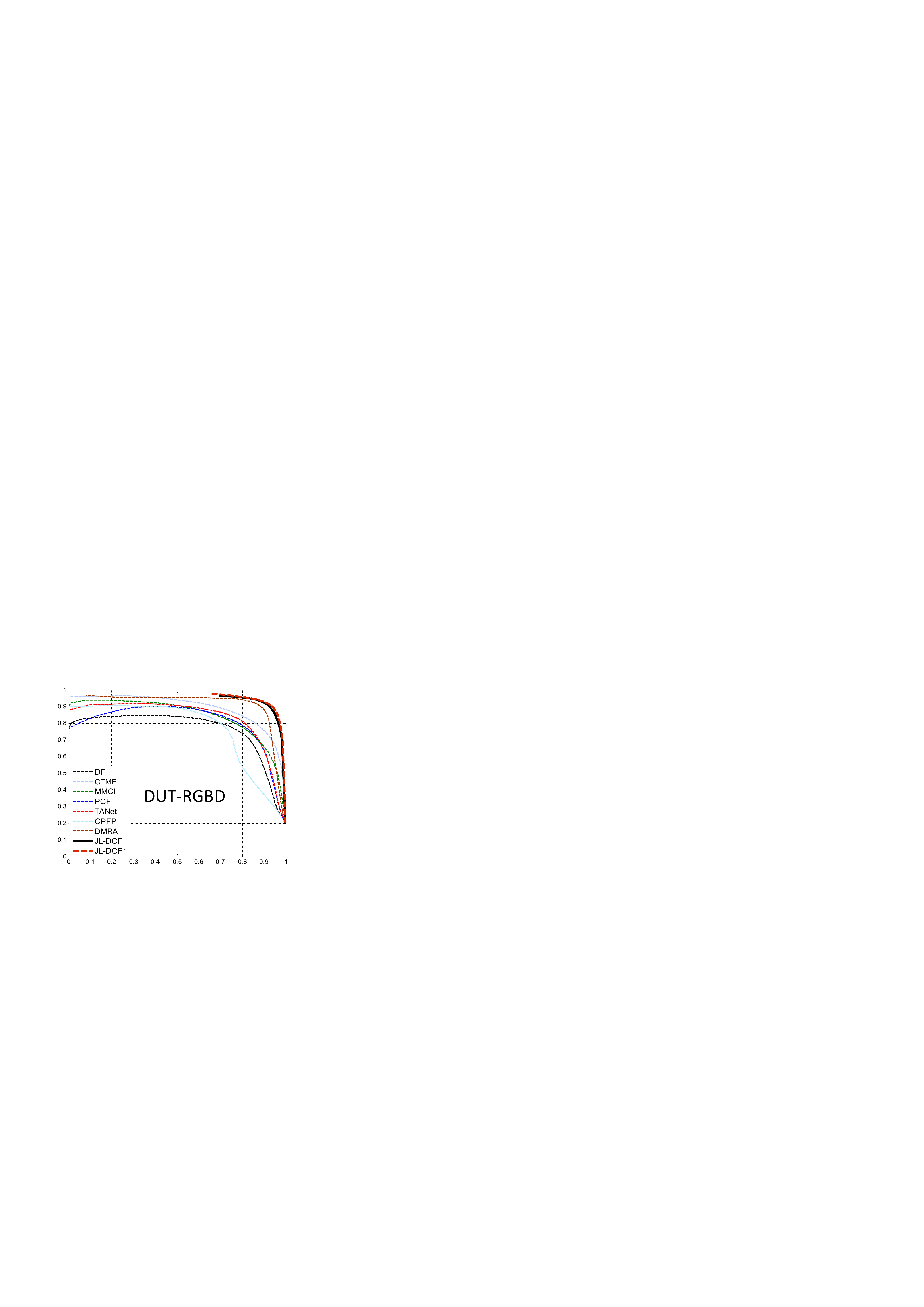}
\vspace{-15pt}
\caption{Precision-recall curves of SOTA methods and the proposed \ourmodel~on DUT-RGBD dataset \cite{Piao2019depth}.}
\label{fig_dutpr}
\end{figure}

We compare \ourmodel~(ResNet configuration) with 14 SOTA methods. Among the competitors, DF~\cite{qu2017rgbd}, AFNet~\cite{wang2019adaptive}, CTMF~\cite{han2017cnns}, MMCI~\cite{chen2019multi}, PCF~\cite{chen2018progressively}, TANet~\cite{chen2019three}, CPFP~\cite{zhao2019contrast}, D3Net~\cite{fan2019rethinking}, and DMRA~\cite{Piao2019depth} are recent deep learning-based methods, while ACSD~\cite{ju2014depth}, LBE~\cite{feng2016local}, DCMC~\cite{cong2016saliency}, MDSF~\cite{song2017depth}, and SE~\cite{guo2016salient} are traditional techniques using various hand-crafted features/hypotheses.
Specifically, ``\ourmodel'' refers to the model obtained using only RGB-D training data, while ``\ourmodel$^*$'' refers to training the model with both RGB-D and RGB data.
Quantitative results on the six widely used datasets are shown in Table \ref{tab_sota}\footnote{There was a small error in the LFSD scores in our previous conference version \cite{fu2020jldcf}, as we later found there was a GT map ``29.png'' corrupted due to format conversion. This error led to a small performance drop for all models, but did not change their relative rankings. We have corrected this GT map as well as all the scores.}. Notable performance gains of \ourmodel~over existing and recently proposed techniques, like CPFP\cite{zhao2019contrast}, D3Net\cite{fan2019rethinking}, and DMRA\cite{Piao2019depth}, can be seen in all four metrics. This validates the consistent effectiveness of \ourmodel~and its generalizability. Besides, as seen in Table \ref{tab_sota},~\ourmodel$^*$ improves the performance over \ourmodel~on most datasets, showing that transferring knowledge from the RGB task to the RGB-D task does benefit the latter and brings solid improvement, \eg, 0.6\% average gain on $S_\alpha$ across all six datasets. Comparisons of precision-recall curves are given in  Fig. \ref{fig_sotapr}, where \ourmodel~and \ourmodel$^*$ achieve the best results compared to all existing techniques.

\begin{figure*}[t!]
\includegraphics[width=1.0\textwidth]{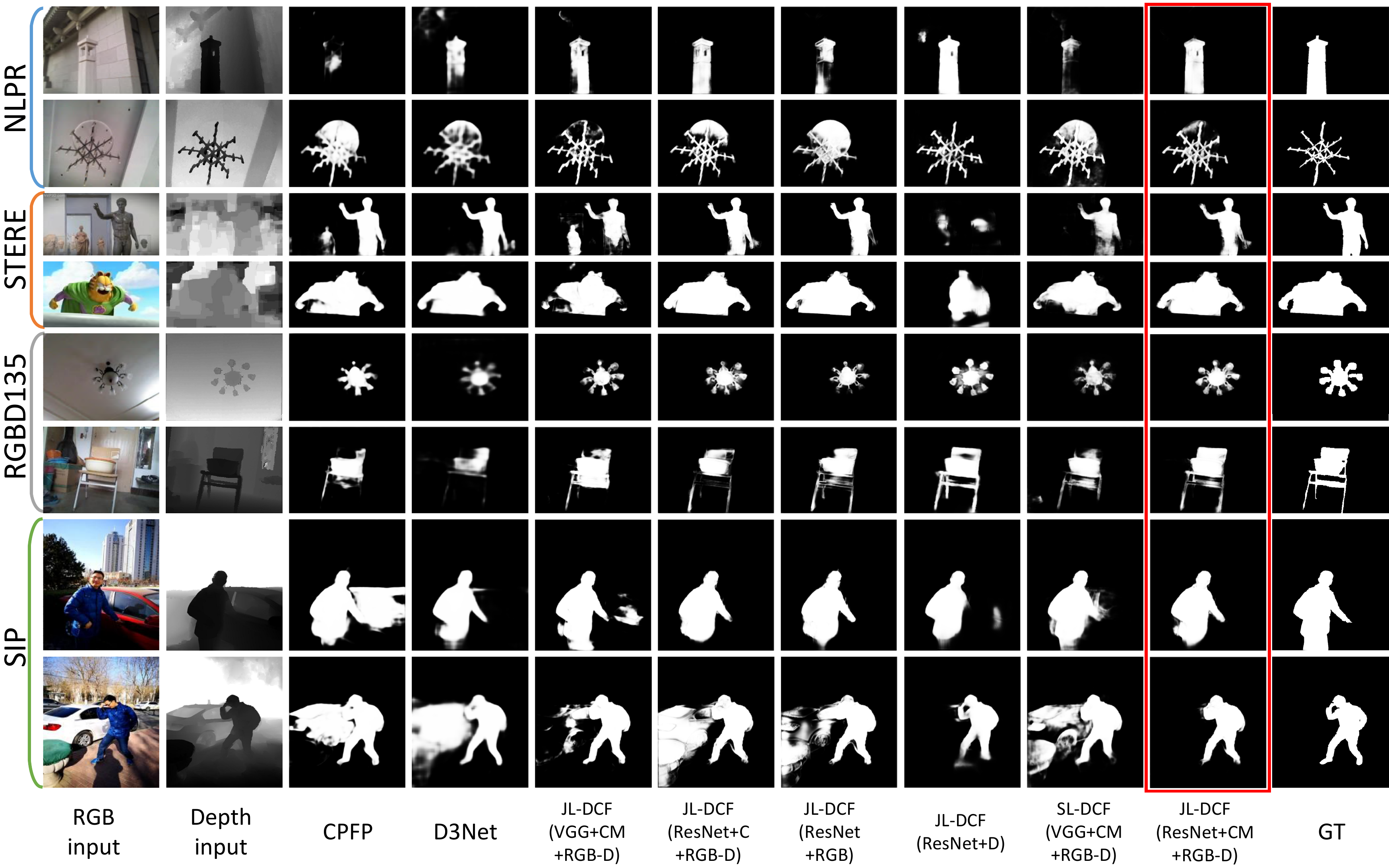}
\vspace{-15pt}
\caption{Visual examples from NLPR, STERE, RGB135, SIP datasets for ablation studies. Generally, the full implementation of \ourmodel~(ResNet+CM+RGB-D, highlighted in the red box) achieves the closest results to the ground truth.}
\label{fig_ablationvisual}
\end{figure*}

Visual examples are shown in Fig. \ref{fig_sotavisual}. \fdp{\ourmodel~and \ourmodel$^*$ appear} to be more effective at utilizing depth information for cross-modal compensation, making it better for detecting target objects in the RGB-D mode. Additionally, the deeply-supervised coarse predictions of \fdp{\ourmodel} are listed in Fig. \ref{fig_sotavisual}. One can see that they provide basic object localization support for the subsequent cross-modal refinement, and our densely cooperative fusion architecture learns an adaptive and ``image-dependent'' way of fusing this support with the hierarchical multi-view features. This proves that the fusion process does not degrade in either of the two views (RGB/depth), leading to boosted performance after fusion.

Table \ref{tab_dutrgbd} and Fig. \ref{fig_dutpr} further show the comparative results on the latest DUT-RGBD dataset \cite{Piao2019depth}. Our \ourmodel~again shows superior performance against all SOTA models. Note that the experimental results on this dataset clearly validate the elegant generalizability of \ourmodel, because it was not trained additionally on the training set of DUT-RGBD which has 800 pairs of RGB and depth images, but still can outperform DMRA, whose training data has included the training set of DUT-RGBD, with notable margins.

\begin{table}[t!]
    \renewcommand{\arraystretch}{0.8}
    \caption{Quantitative measures on the DUT-RGBD testing set (400 images) \cite{Piao2019depth}. Compared models are those whose results on this dataset are publicly available and include: DF \cite{qu2017rgbd}, CTMF \cite{han2017cnns}, MMCI \cite{chen2019multi}, PCF \cite{chen2018progressively}, TANet \cite{chen2019three}, CPFP \cite{zhao2019contrast}, DMRA \cite{Piao2019depth}, \ourmodel~(Ours) and \ourmodel$^*$ (Ours$^*$).
    }\label{tab_dutrgbd}
    \centering
    \footnotesize
    \setlength{\tabcolsep}{0.95mm}
    \begin{tabular}{r||c|c|c|c|c|c|c|c|c}
    \hline
          Metric  & \tabincell{c}{\cite{qu2017rgbd}} & \tabincell{c}{\cite{han2017cnns}} & \tabincell{c}{\cite{chen2019multi}} & \tabincell{c}{\cite{chen2018progressively}} & \tabincell{c}{\cite{chen2019three}} & \tabincell{c}{\cite{zhao2019contrast}} & \tabincell{c}{\cite{Piao2019depth}}& \tabincell{c}{Ours}&\tabincell{c}{Ours$^*$}\\
    \hline
    \hline
           $S_\alpha\uparrow$ &0.730&0.831&0.791&0.801 &0.808&0.749&0.889&\emph{0.905}&\textbf{0.913}\\
           $F_{\beta}^{\textrm{max}}\uparrow$  &0.734&0.823& 0.767 & 0.771& 0.790 & 0.718& 0.898 & \emph{0.911} & \textbf{0.916}\\
           $E_{\phi}^{\textrm{max}}\uparrow$   & 0.819 & 0.899& 0.859& 0.856& 0.861 & 0.811& 0.933 & \emph{0.943} & \textbf{0.949}\\
           $M\downarrow$      & 0.145 & 0.097& 0.113& 0.100& 0.093 & 0.099& 0.048 & \emph{0.042} & \textbf{0.039}\\
    \hline
\end{tabular}
\end{table}

\subsection{Ablation Studies}\label{sec44}
We conduct thorough ablation studies by removing or replacing components from the full implementation of \ourmodel. We set the ResNet version of \ourmodel~(trained with only RGB-D data) as reference, and then compare various ablated/modified models to it. We denote this reference version as ``\ourmodel~(ResNet+CM+RGB-D)'', where ``CM'' refers to the usage of CM modules and ``RGB-D'' refers to both RGB and depth inputs.

Firstly, to compare different backbones, a version ``\ourmodel~(VGG+CM+RGB-D)'' is trained by replacing the ResNet backbone with VGG, while keeping other settings unchanged. To validate the effectiveness of the adopted cooperative fusion modules, we train another version ``\ourmodel~(ResNet+C+RGB-D)'', by replacing the CM modules with a concatenation operation. To demonstrate the effectiveness of combining RGB and depth, we train two versions ``\ourmodel~(ResNet+RGB)'' and ``\ourmodel~(ResNet+D)'' respectively, where all the batch-related operations (such as CM modules) in Fig. \ref{fig_blockdiagram} are replaced with identity mappings, while all the other settings, including the dense decoder and deep supervision, are kept unchanged. Note that this validation is important to show that our network has learned complementary information by fusing RGB and depth. Lastly, to illustrate the benefit of joint learning, we train a scheme ``SL-DCF (VGG+CM+RGB-D)'' using two separate backbones for RGB and depth. ``SL'' stands for ``Separate Learning'', in contrast to the proposed ``Joint Learning''. In this test, we adopt VGG-16, which is smaller, since using two separate backbones leads to almost twice the overall model size.

Quantitative comparisons for various metrics are shown in Table \ref{tab_ablation}. Two SOTA methods, CPFP \cite{zhao2019contrast} and D3Net \cite{fan2019rethinking}, are listed for reference. Fig. \ref{fig_ablationvisual} shows visual ablation comparisons. Five different observations can be made:

\textbf{I. ResNet-101 $vs.$ VGG-16:} From the comparison between columns ``\texttt{A}'' and ``\texttt{B}'' in Table \ref{tab_ablation}, the superiority of the ResNet backbone over VGG-16 is evident, which is consistent with previous works. Note that the VGG version of our scheme still outperforms the leading methods CPFP (VGG-16 backbone) and D3Net (ResNet backbone).

\textbf{II. Effectiveness of CM modules:} Comparing columns ``\texttt{A}'' and ``\texttt{C}''  demonstrates that changing the CM modules into concatenation operations leads to a certain amount of degeneration. The underlying reason is that the whole network tends to bias its learning towards only RGB information, while ignoring depth, since it is able to achieve fairly good results (column ``\texttt{D}'') by doing so on most datasets. Although concatenation is a popular way to fuse features, the learning may become easily trapped without appropriate guidance. In contrast, our CM modules perform the ``explicit fusion operation'' across RGB and depth modalities.

\begin{figure}
\centering
\includegraphics[width=0.40\textwidth]{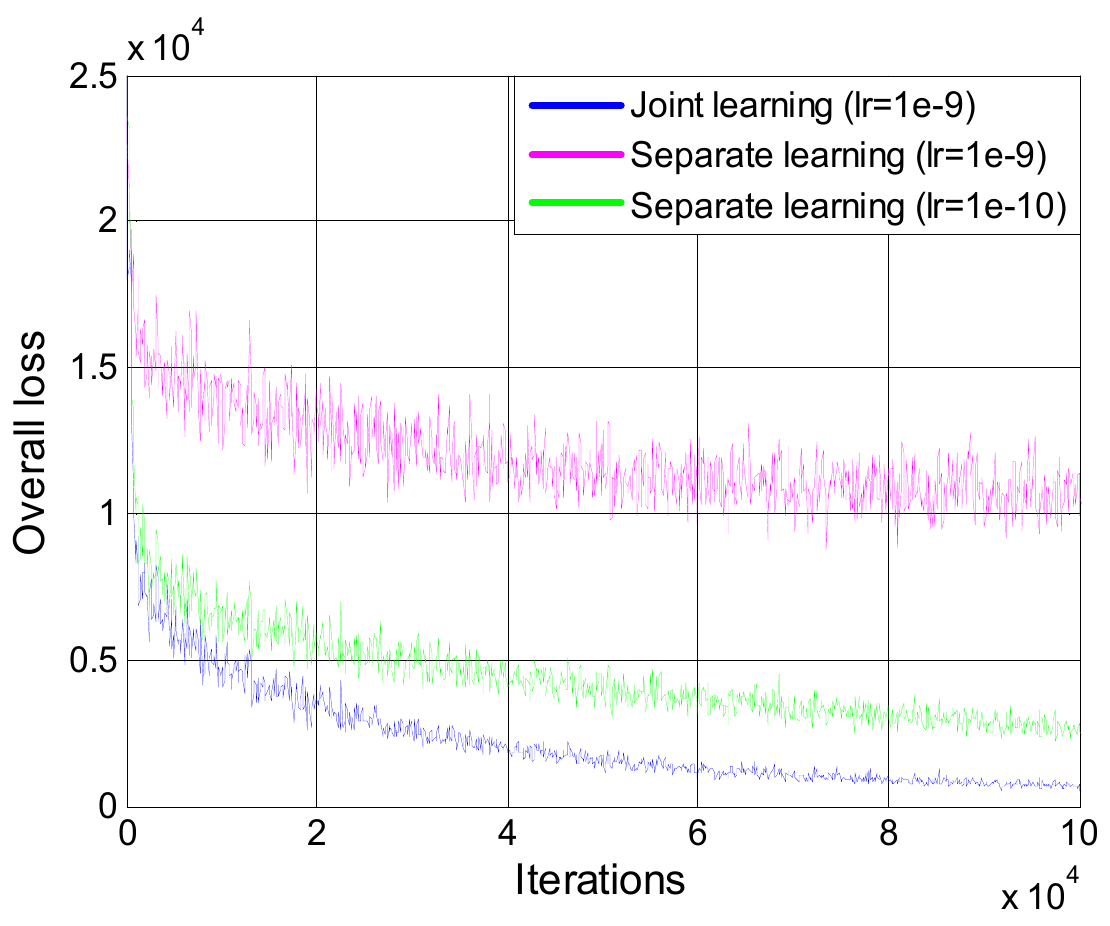}
\vspace{-10pt}
\caption{Learning curve comparison between joint learning (\ourmodel) and separate learning (SL-DCF).}
\label{fig_learningcurve}
\vspace{-0.3cm}
\end{figure}

\textbf{III. Combining RGB and depth:} The effectiveness of combining RGB and depth for boosting the performance is clearly validated by the consistent improvement over most datasets (compare column ``\texttt{A}''  with columns ``\texttt{D}'' and ``\texttt{E}''). The only exception is on STERE \cite{niu2012leveraging}, with the reason being that the quality of depth maps in this dataset is much worse compared to other datasets. Visual examples are shown in Fig. \ref{fig_ablationvisual}, in the 3$^{rd}$ and 4$^{th}$ rows. We find that many depth maps from STERE are too coarse and have very inaccurate object boundaries, misaligning with the true objects. Absorbing such unreliable depth information may, in turn, degrade the performance. Quantitative evidence can be seen in Table \ref{tab_ablation}, column ``\texttt{E}'' (STERE dataset), where solely using depth cues achieves much worse performance (about 16\%/20\% lower on $S_{\alpha}$/$F_{\beta}^{\textrm{max}}$ compared to RGB) than on other datasets.

\textbf{IV. RGB only $vs.$ depth only:} The comparison between columns ``\texttt{D}'' and ``\texttt{E}'' in Table \ref{tab_ablation} proves that using RGB data for saliency estimation is superior to using depth in most cases, indicating that the RGB view is generally more informative. However, using depth information achieves better results than RGB on SIP \cite{fan2019rethinking} and RGBD135 \cite{cheng2014depth}, as visualized in Fig. \ref{fig_ablationvisual}. This implies that the depth maps from the two datasets are of relatively good quality.

\textbf{V. Efficiency of JL component:}
Existing models usually use separate learning approaches to extract features from RGB and depth data, respectively. In contrast, our \ourmodel~adopts a joint learning strategy to obtain the features simultaneously.
We compare the two learning strategies and
%
find that using separate learning (two separate backbones) is likely to increase the training difficulties.
Fig. \ref{fig_learningcurve} shows typical learning curves for such a case. In the separate learning setting, where the initial learning rate is $lr=10^{-9}$, the network is easily trapped in a local optimum with high loss, while the joint learning setting (shared network) can converge nicely.
Further, for separate learning, if the learning rate is set to $lr=10^{-10}$, the learning process is rescued from local oscillation but converges slowly compared to our joint learning strategy. As shown in columns ``\texttt{B}'' and ``\texttt{F}'' in Table \ref{tab_ablation}, the resulting converged model after 40 epochs achieves worse performance than \ourmodel, namely 1.1\%/1.76\% overall drop on $S_{\alpha}$/$F_{\beta}^{\textrm{max}}$. We attribute the better performance of \ourmodel~to its joint learning from both RGB and depth data.

\begin{table}[t!]
    \renewcommand{\arraystretch}{0.8}
    \caption{Quantitative evaluation for ablation studies described in Section \ref{sec44}. For different configurations, ``\texttt{A}'': JL-DCF (ResNet+CM+RGB-D), ``\texttt{B}'': JL-DCF (VGG+CM+RGB-D), ``\texttt{C}'': JL-DCF (ResNet+C+RGB-D), ``\texttt{D}'': JL-DCF (ResNet+RGB), ``\texttt{E}'': JL-DCF (ResNet+D), ``\texttt{F}'': SL-DCF (VGG+CM+RGB-D).
    }\label{tab_ablation}
    \centering
    \footnotesize
    \setlength{\tabcolsep}{1.0mm}
    \begin{tabular}{p{0.8mm}p{0.8mm}r||c|c|c|c|c|c|c|c}
    \hline
          && Metric  & \tabincell{c}{CPFP} & \tabincell{c}{D3Net} & \tabincell{c}{\texttt{A}}  & \tabincell{c}{\texttt{B}} & \tabincell{c}{\texttt{C}} & \tabincell{c}{\texttt{D}} & \tabincell{c}{\texttt{E}} & \tabincell{c}{\texttt{F}}\\
    \hline
    \hline
          \multirow{4}{*}{\begin{sideways}\textit{NJU2K}\end{sideways}} & \multirow{4}{*}{\begin{sideways}\cite{ju2014depth}\end{sideways}}
          & $S_\alpha\uparrow$ &0.878&0.895&\textbf{0.903}&0.897
&0.900&0.895&0.865&0.886\\
                                                                        && $F_{\beta}^{\textrm{max}}\uparrow$       &0.877&0.889
 & \textbf{0.903} & 0.899& 0.898 & 0.892& 0.863 & 0.883\\
                                                                        && $E_{\phi}^{\textrm{max}}\uparrow$       & 0.926 & 0.932& \textbf{0.944}& 0.939& 0.937 & 0.937& 0.916 & 0.929\\
                                                                        && $M\downarrow$       & 0.053 & 0.051& \textbf{0.043}& 0.044& 0.045 & 0.046& 0.063 & 0.053\\
    \hline
        \multirow{4}{*}{\begin{sideways}\textit{NLPR}\end{sideways}} & \multirow{4}{*}{\begin{sideways}\cite{peng2014rgbd}\end{sideways}}& $S_\alpha\uparrow$       & 0.888 & 0.906& \textbf{0.925}& 0.920& 0.924 & 0.922& 0.873 & 0.901\\
                                                                        && $F_{\beta}^{\textrm{max}}\uparrow$       & 0.868& 0.885& \textbf{0.916}& 0.907& 0.914 & 0.909& 0.843 & 0.881\\
                                                                        && $E_{\phi}^{\textrm{max}}\uparrow$       & 0.932 & 0.946& \textbf{0.962}& 0.959& 0.961 & 0.957& 0.930 & 0.946\\
                                                                        && $M\downarrow$       & 0.036 & 0.034& \textbf{0.022}& 0.026& 0.023 & 0.025& 0.041 & 0.033\\
    \hline
        \multirow{4}{*}{\begin{sideways}\textit{STERE}\end{sideways}}& \multirow{4}{*}{\begin{sideways}\cite{niu2012leveraging}\end{sideways}} & $S_\alpha\uparrow$       & 0.879& 0.891& 0.905& 0.894& 0.906 & \textbf{0.909}& 0.744 & 0.886\\
                                                                        && $F_{\beta}^{\textrm{max}}\uparrow$       & 0.874& 0.881& \textbf{0.901}& 0.889& 0.899& 0.901 & 0.708& 0.876\\
                                                                        && $E_{\phi}^{\textrm{max}}\uparrow$       & 0.925 & 0.930& \textbf{0.946}& 0.938& 0.945 & 0.946& 0.834 & 0.931\\
                                                                        && $M\downarrow$       & 0.051 & 0.054& 0.042& 0.046& 0.041 & \textbf{0.038}& 0.110 & 0.053\\
    \hline
        \multirow{4}{*}{\begin{sideways}\textit{RGBD135}\end{sideways}}& \multirow{4}{*}{\begin{sideways}\cite{cheng2014depth}\end{sideways}}   & $S_\alpha\uparrow$       & 0.872& 0.904& \textbf{0.929}& 0.913& 0.916 & 0.903& 0.918 & 0.893\\
                                                                        && $F_{\beta}^{\textrm{max}}\uparrow$       & 0.846& 0.885& \textbf{0.919}& 0.905& 0.906 & 0.894& 0.906 & 0.876\\
                                                                        && $E_{\phi}^{\textrm{max}}\uparrow$    & 0.923 & 0.946& \textbf{0.968}& 0.955& 0.957 & 0.947& 0.967 & 0.950\\
                                                                        && $M\downarrow$       & 0.038 & 0.030& \textbf{0.022}& 0.026& 0.025 & 0.027& 0.027 & 0.033\\
                                           \hline
        \multirow{4}{*}{\begin{sideways}\textit{LFSD}\end{sideways}}& \multirow{4}{*}{\begin{sideways}\cite{li2014saliency}\end{sideways}}   & $S_\alpha\uparrow$       & 0.820& 0.832& \textbf{0.862}& 0.841& 0.860 & 0.853& 0.760 & 0.834\\
                                                                        && $F_{\beta}^{\textrm{max}}\uparrow$       & 0.821& 0.819& \textbf{0.866}& 0.844& 0.858 & 0.850& 0.768 & 0.832\\
                                                                        && $E_{\phi}^{\textrm{max}}\uparrow$       & 0.864 & 0.864 & \textbf{0.901}& 0.885& 0.901 & 0.897& 0.824 & 0.872\\
                                                                        && $M\downarrow$       & 0.095 & 0.099& \textbf{0.071}& 0.084& 0.071 & 0.076& 0.119 & 0.093\\
     \hline
        \multirow{4}{*}{\begin{sideways}\textit{SIP}\end{sideways}}& \multirow{4}{*}{\begin{sideways}\cite{fan2019rethinking}\end{sideways}} & $S_\alpha\uparrow$       & 0.850& 0.864& \textbf{0.879}& 0.866& 0.870 & 0.855& 0.872 & 0.865\\
                                                                        && $F_{\beta}^{\textrm{max}}\uparrow$       & 0.851& 0.862& \textbf{0.885}& 0.873& 0.873 & 0.857& 0.877 & 0.863\\
                                                                        && $E_{\phi}^{\textrm{max}}\uparrow$       & 0.903 & 0.910& \textbf{0.923}& 0.916& 0.916 & 0.908& 0.920 & 0.913
\\                                                                     && $M\downarrow$   & 0.064 & 0.063& \textbf{0.051}& 0.056& 0.055& 0.061& 0.056 & 0.061\\
\hline
\end{tabular}
\end{table}

\fkr{
\textbf{Further ablation analyses:}
Besides the five key observations above, there are also other flexible parts in \ourmodel~to discuss, such as the FA modules and dense connections. Consequently, we have formed extra configurations ``\texttt{G}''$\sim$``\texttt{J}'', where
``\texttt{G}'': removing all FA modules from ``\texttt{A}\footnote{It indicates the aforementioned ``\texttt{A}'' in Table \ref{tab_ablation}, and similarly hereinafter.}'' to get a degenerated decoder which linearly sums up skips from all scales;
``\texttt{H}'': removing all dense connections from  ``\texttt{A}''; ``\texttt{I}'': removing all dense connections from ``\texttt{A}'', while leaving only the skip connection from FA5 to FA1, as a residual way;
``\texttt{J}'': replacing the ResNet-101 backbone with a more powerful DenseNet-161 \cite{huang2017densely} to show whether potential boost of \ourmodel~can be obtained by other advanced backbones. For ``\texttt{J}'', the DenseNet is incorporated into \ourmodel~by connecting \emph{side} \emph{path1}$\sim$\emph{path6} to \emph{conv1\_2} of VGG-16 (similar to the ResNet configuration in Section \ref{sec42}), and \emph{conv0}, \emph{denseblock1}$\sim$\emph{denseblock4} of the DenseNet.
}

\fkr{
Results are shown in Table \ref{tab_ablation2}. In brief, we find that adding FA modules (\ie, ``\texttt{A}'') for non-linear aggregation makes the network more powerful, while removing all FA modules (\ie, ``\texttt{G}'') results in average $\sim$1.38\% $F_{\beta}^{\textrm{max}}$ drop. 
Regarding the employed dense connections, one can see that ``\texttt{A}'' achieves improvement over ``\texttt{H}'' on most datasets (except on RGBD135 where similar results are obtained), showing that dense connections could somewhat enhance robustness of the network. Another interesting observation is that the residual connection ``\texttt{I}'' works comparably well on NJU2K, STERE and RGBD135. This is because although the residual connection is simplified from the dense connections, it alleviates the gradual dilution of deep location information and offers extra high-level guidance, as also observed in \cite{liu2019simple}.
About ``\texttt{J}'', we witness very encouraging boost by further switching the backbone from ResNet-101 to DenseNet-161. This indicates that more powerful backbones are able to play their roles in our \ourmodel~framework.
}


\begin{table}[t!]
\centering
\caption{Further ablation analyses, where details about ``\texttt{G}''$\sim$``\texttt{J}'' can be found in Section \ref{sec44}. Here $F_{\beta}$ means $F_{\beta}^{\textrm{max}}$, whose superscript is omitted for the sake of space. }\label{tab_ablation2}
\footnotesize
{
\setlength{\tabcolsep}{0.3mm}
\begin{tabular}{c||c|c|c|c|c|c|c|c|c|c|c|c}

\hline
             & \multicolumn{2}{c|}{\textit{NJU2K}}  & \multicolumn{2}{c|}{\textit{NLPR}}  & \multicolumn{2}{c|}{\textit{STERE}}  & \multicolumn{2}{c|}{\textit{RGBD135}}  & \multicolumn{2}{c|}{\textit{LFSD}} & \multicolumn{2}{c}{\textit{SIP}}\\
\hline
             & $S_{\alpha}\uparrow$ & $F_{\beta}\uparrow$   & $S_{\alpha}\uparrow$ & $F_{\beta}\uparrow$  & $S_{\alpha}\uparrow$ & $F_{\beta}\uparrow$  & $S_{\alpha}\uparrow$ & $F_{\beta}\uparrow$  & $S_{\alpha}\uparrow$ & $F_{\beta}\uparrow$  & $S_{\alpha}\uparrow$ & $F_{\beta}\uparrow$\\
\hline
\hline
\texttt{A}       & 0.903  & 0.903 & 0.925 & 0.916  & 0.905 & 0.901 & 0.929  & 0.919 & 0.862 & 0.866  & 0.879 & 0.885\\
\hline
\texttt{G}        &  0.893  & 0.893 & 0.911 & 0.894  & 0.893 & 0.884 & 0.924  & 0.912 & 0.855 & 0.852  & 0.870 & 0.872\\
\hline
\texttt{H}        & 0.902  & 0.902 & 0.922 & 0.911  & 0.904 & 0.898 & 0.930  & 0.923 & 0.854 & 0.857  & 0.874 & 0.879\\
\hline
\texttt{I}        & 0.904  & 0.906 & 0.924 & 0.913  & 0.905 & 0.901 & 0.929  & 0.921 & 0.859 & 0.861  & 0.876 & 0.881\\
\hline
\texttt{J}        &  \textbf{0.917}  & \textbf{0.917} & \textbf{0.934} & \textbf{0.924}  & \textbf{0.909} & \textbf{0.905} & \textbf{0.934}  & \textbf{0.926} & \textbf{0.863} & \textbf{0.868}  & \textbf{0.894} & \textbf{0.903}\\
\hline
\end{tabular}
}
\end{table}

\subsection{Computational Efficiency}\label{sec45}

We evaluate the computation time of \ourmodel~on a desktop equipped with an Intel I7-8700K CPU (3.7GHz), 16G RAM, and NVIDIA 1080Ti GPU. \ourmodel~ is implemented on Caffe \cite{jia2014caffe}. We test the inference time of our models using the Matlab interface of Caffe, over the 100 samples (resized to $320 \times 320$) from the LFSD dataset. The average GPU inference times are given in Table \ref{tab_time}.

\begin{table}[!htb]
\centering
\caption{Average GPU inference times (second) of \ourmodel.}\label{tab_time}
{
\renewcommand{\tabcolsep}{4mm}
\begin{tabular}{c||c||c||c}
\hline
Backbones\textbackslash Components     & Overall & JL & DCF \\
\hline
\hline
VGG-16        & 0.089  & 0.065 & 0.024\\
ResNet-101        &  0.111 & 0.087 & 0.024\\
\hline
\end{tabular}
}
\end{table}

As can be seen, the JL (joint learning) component in \ourmodel, which includes the shared backbone, consumes most of the time, while the DCF (densely cooperative fusion) component takes only 0.024s. \fdp{Note that the latter fact actually implies that our introduced CM and FA modules, as well as dense connections, result in only little computation load, since the entire DCF component is generally efficient. For example, the extra dense connections lead to only 0.008s gain.} Besides, the ResNet-101 is computationally 0.022s slower than VGG-16 due to its higher number of network parameters. This reveals that, in \ourmodel, the backbone dominates the time cost, and a way for acceleration is to utilize a light-weighted backbone; however the impact on detection accuracy should be considered at the same time.

\subsection{Application to Other Multi-modal Fusion Tasks}\label{sec46}

\begin{figure}[thbp]
\centering
\includegraphics[width=0.40\textwidth]{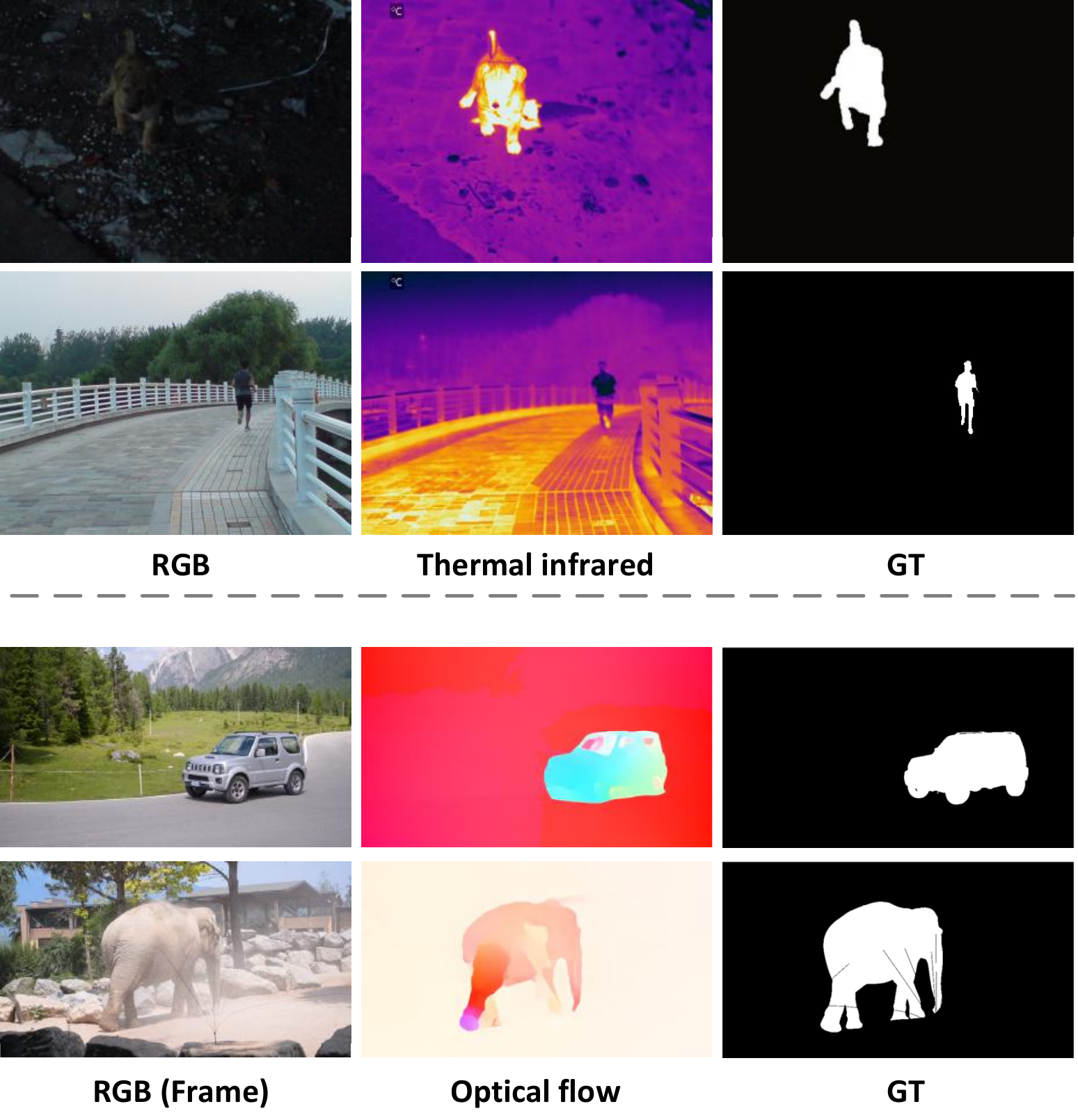}
\vspace{-10pt}
\caption{Illustration of the commonality and complementarity of thermal infrared images (upper two rows) and optical flow images (lower two rows) to the RGB ones for SOD. Complementary to the RGB view, salient objects sometimes are easier to distinguish in these two views. Meanwhile, salient objects ``stand out'' in these two views like they do in the RGB view, indicating certain commonality.}
\label{fig_multimoda}
\end{figure}

Although the proposed \ourmodel~is originally motivated by and evaluated on the RGB-D SOD task, thanks to its general design for exploiting cross-modal commonality and complementarity, it can be applied to other closely-related multi-modal SOD tasks, such as RGB-T (``T'' refers to thermal infrared) SOD \cite{tu2020multi,tu2019rgb,tang2019rgbt,zhang2019rgb} and video SOD (VSOD) \cite{wang2017video,li2018flow,li2018unsupervised,song2018pyramid,fan2019shifting,gu2020pyramid}. Intuitively, salient objects can present similar saliency characters in thermal infrared images (Fig. \ref{fig_multimoda} upper part) and optical flow images (Fig. \ref{fig_multimoda} lower part) as they generally present in RGB images. Therefore, for SOD, there exists certain commonality between thermal/flow images and RGB images, as indicated by many traditional models \cite{cong2019video,xu2019video,xu2019video2} that are based on hand-crafted features. Examples for explaining this concept are shown in Fig. \ref{fig_multimoda}. To apply \ourmodel~to RGB-T SOD and VSOD, we just change the training data of \ourmodel~from paired RGB and depth data to paired RGB and thermal/flow data, without any other modification to the framework. In addition, because the thermal and flow images are commonly converted to the three-channel RGB format, applying a Siamese network to RGB \emph{vs.} thermal/flow is straightforward.

\textbf{RGB-T (thermal infrared) SOD.} Since, to date, there are only a small number of works related to RGB-T SOD \cite{tu2020multi,tu2019rgb,tang2019rgbt,zhang2019rgb}, there lack universally-agreed evaluation protocols and benchmarks. Following a recent work \cite{tu2020multi}, we test \ourmodel~on VT821 \cite{tang2019rgbt}, which is an RGB-T SOD dataset having 821 samples of aligned RGB and thermal images, and compare our results with those provided by the authors of \cite{tu2020multi}. VT1000 \cite{tu2019rgb}, which contains 1000 samples, is adopted as the training set. The method proposed in \cite{tu2020multi} is referred to as MIED. Meanwhile, \cite{tu2020multi} also provides us the adapted results of DMRA \cite{Piao2019depth}, retrained on VT1000.

Quantitative evaluation results on VT821 are shown in Table \ref{tab_vt821}, where we report four different versions of \ourmodel: ``\ourmodel'', ``\ourmodel(T)'', ``\ourmodel$^*$'', ``\ourmodel$^*$(T)''. ``\ourmodel'' and ``\ourmodel$^*$'' are the same models tested in Tab. \ref{tab_sota}, trained on the RGB-D SOD task. ``\ourmodel(T)'' and ``\ourmodel$^*$(T)'' refer to the \ourmodel~models retrained on the RGB-T data, \ie, VT1000 (40 epochs, initialized consistently as the RGB-D task), where the latter means training the model jointly with both RGB-T and RGB data, similar to the RGB-D case mentioned before. From Table \ref{tab_vt821}, first, it can be seen that our four models outperform MIED and DMRA consistently on the two metrics. Surprisingly, even the models (\eg, \ourmodel, \ourmodel$^*$) trained by RGB-D data can generalize well to this RGB-T SOD task, further validating the robustness and generalizability of our framework. Still, co-training with more RGB data enhances the detection accuracy, whereas re-training \ourmodel~by RGB-T data better adapts it to the specific task. Undoubtedly, the best performance is attained by model ``\ourmodel$^*$(T)'', surpassing MIED by 2.6\% on $S_{\alpha}$.

\begin{table}[t!]
    \renewcommand{\arraystretch}{0.8}
    \caption{Comparing \ourmodel~to existing RGB-T SOD models on VT821~\cite{tang2019rgbt} dataset.
    }\label{tab_vt821}
    \centering
    \footnotesize
    \setlength{\tabcolsep}{1.05mm}
    \begin{tabular}{r||c|c|c|c|c|c}
    \hline
          Metric  & MIED & DMRA & \ourmodel & \ourmodel(T) & \ourmodel$^*$ & \ourmodel$^*$(T)\\
    \hline
    \hline
           $S_\alpha\uparrow$ & 0.866 & 0.844 & 0.873 & 0.876 & \emph{0.885} & \textbf{0.892}\\
          $M\downarrow$      & 0.053 & 0.049 & 0.037 & 0.037 & \textbf{0.031} & \emph{0.033}\\
    \hline
\end{tabular}
\end{table}

\begin{table}[t]
\renewcommand{\arraystretch}{0.8}
\caption{Comparing \ourmodel~to existing VSOD models on five widely used benchmark datasets.
}\label{tab_vsod}
\centering
\footnotesize
{
\setlength{\tabcolsep}{0.5mm}
\begin{tabular}{r|cc|cc|cc|cc|cc}

\hline
             & \multicolumn{2}{c|}{\textit{DAVIS-T}}  & \multicolumn{2}{c|}{\textit{FBMS-T}}  & \multicolumn{2}{c|}{\textit{ViSal}}  & \multicolumn{2}{c|}{\textit{VOS}}  & \multicolumn{2}{c}{\textit{DAVSOD}} \\

             & \multicolumn{2}{c|}{\cite{perazzi2016benchmark}}  & \multicolumn{2}{c|}{\cite{ochs2013segmentation}}  & \multicolumn{2}{c|}{\cite{wang2015consistent}}  & \multicolumn{2}{c|}{\cite{li2017benchmark}}  & \multicolumn{2}{c}{\cite{fan2019shifting}} \\
\cline{2-11}
             Model & $S_{\alpha}\uparrow$ & $M\downarrow$   & $S_{\alpha}\uparrow$ & $M\downarrow$  & $S_{\alpha}\uparrow$ & $M\downarrow$  & $S_{\alpha}\uparrow$ & $M\downarrow$  & $S_{\alpha}\uparrow$ & $M\downarrow$ \\
\hline
\hline
\scriptsize{DLVS \cite{wang2017video}}       & 0.794  & 0.061 & 0.794 & 0.091  & 0.881 & 0.048 & 0.760  & 0.099 & 0.657 & 0.129  \\
\scriptsize{FGRN \cite{li2018flow}}     & 0.838  & 0.043 & 0.809 & 0.088  & 0.861 & 0.045 & 0.715  & 0.097 & 0.693 & 0.098  \\
\scriptsize{MBNM \cite{li2018unsupervised}}    & 0.887  & 0.031 & 0.857 & 0.047 & 0.898 & 0.020 & 0.742  & 0.099 & 0.637 & 0.159 \\
\scriptsize{PDBM \cite{song2018pyramid}}     & 0.882  & 0.028 & 0.851 & 0.064  & 0.907 & 0.032 & 0.818  & 0.078 & 0.698 & 0.116 \\
\scriptsize{SSAV  \cite{fan2019shifting}}     & 0.893  & 0.028 & 0.879 & \textbf{0.040}  & 0.943 & 0.020 & 0.819  & 0.073 & 0.724 & 0.092 \\
\scriptsize{PCSA  \cite{gu2020pyramid}}   & 0.902  & 0.022 & 0.866 & 0.041  & \textbf{0.946} & 0.017 & \textbf{0.827}  & 0.065 & 0.741 & \textbf{0.086} \\
\hline
\ourmodel$^*$ &  \textbf{0.903}  & \textbf{0.022} & \textbf{0.884} & 0.044  & 0.940 & \textbf{0.017} & 0.825  & \textbf{0.063} & \textbf{0.756} & 0.091  \\
\hline
\end{tabular}
}
\end{table}

\begin{figure*}
\includegraphics[width=1\textwidth]{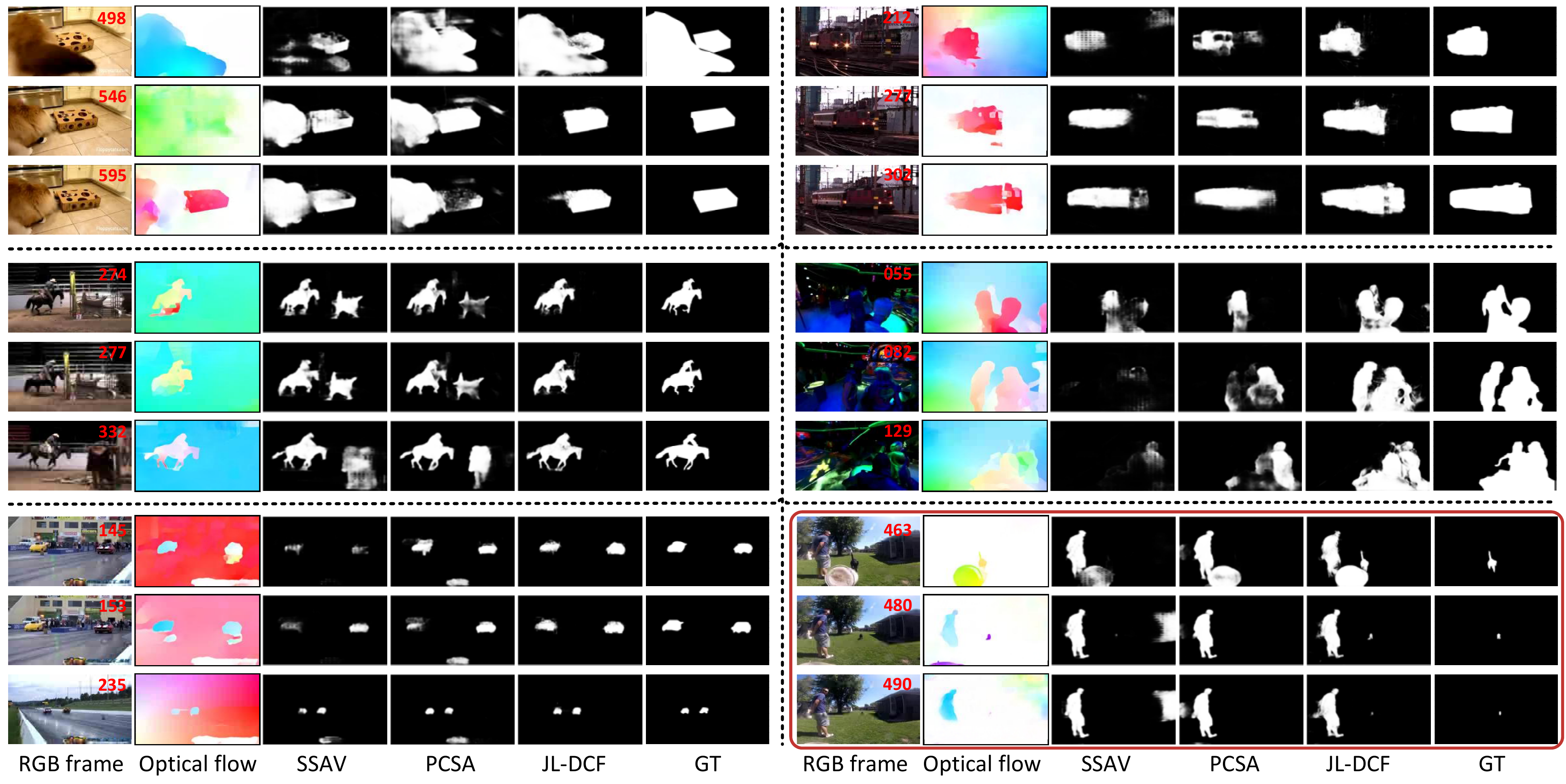}
\vspace{-15pt}
\caption{Visual comparisons of \ourmodel~with two latest SOTA VSOD models: SSAV-CVPR19 \cite{fan2019shifting} and PCSA-AAAI20 \cite{gu2020pyramid}. The bottom right group of images show a failure case where distracting objects are detected by all models. However, note that only \ourmodel~gives responses to the small target dog.}
\label{fig_vsod}
\end{figure*}

\textbf{Video SOD.} \ourmodel~can also be applied to VSOD. We first compute  forward optical flow maps of RGB frames using FlowNet 2.0 \cite{ilg2017flownet}, which is a SOTA deep model for optical flow estimation. A computed flow map originally has two channels for indicating motion displacements. To input it to the JL component of \ourmodel, we convert it to a three-channel color map by using the common flow-field color coding technique \cite{ilg2017flownet}. We train \ourmodel~using the official training sets of DAVIS (30 clips) \cite{perazzi2016benchmark} and FBMS (29 clips) \cite{ochs2013segmentation}, resulting in a total of 2373 samples of paired RGB and flow images. Besides, we find that in this task, co-training with RGB data is essential to the generalization of the model\footnote{Note that most of the existing deep-based VSOD works adopt additional RGB SOD data during their training, such as \cite{fan2019shifting,gu2020pyramid,song2018pyramid}.}, because the scene diversity of the training samples is quite limited\footnote{Most samples are consecutive frames with the same objects with similar background.}. Following \cite{fan2019shifting}, evaluation is conducted on five widely used benchmark datasets: FBMS-T \cite{ochs2013segmentation} (30 clips), DAVIS-T \cite{perazzi2016benchmark} (20 clips), ViSal \cite{wang2015consistent} (17 clips), MCL \cite{kim2015spatiotemporal} (9 clips), UVSD \cite{liu2016saliency} (18 clips), VOS \cite{li2017benchmark} (40 clips selected by \cite{fan2019shifting}), DAVSOD \cite{fan2019shifting} (the easy subset with 35 clips).
As can be seen from Tab. \ref{tab_vsod}, although \ourmodel~is not specially designed for VSOD (no any long-term temporal consideration \cite{fan2019shifting,gu2020pyramid,song2018pyramid}), it is able to achieve comparable performance against SOTAs by learning from RGB and motion images, achieving the best on six out of the ten scores. This, again, shows that \ourmodel~may become a unified and general framework for solving multi-modal feature learning and fusion problems, as it is the first work to exploit both the cross-modal commonality and complementarity. Fig. \ref{fig_vsod} shows several visual comparisons.

\begin{figure}[t!]
\centering
\includegraphics[width=0.46\textwidth]{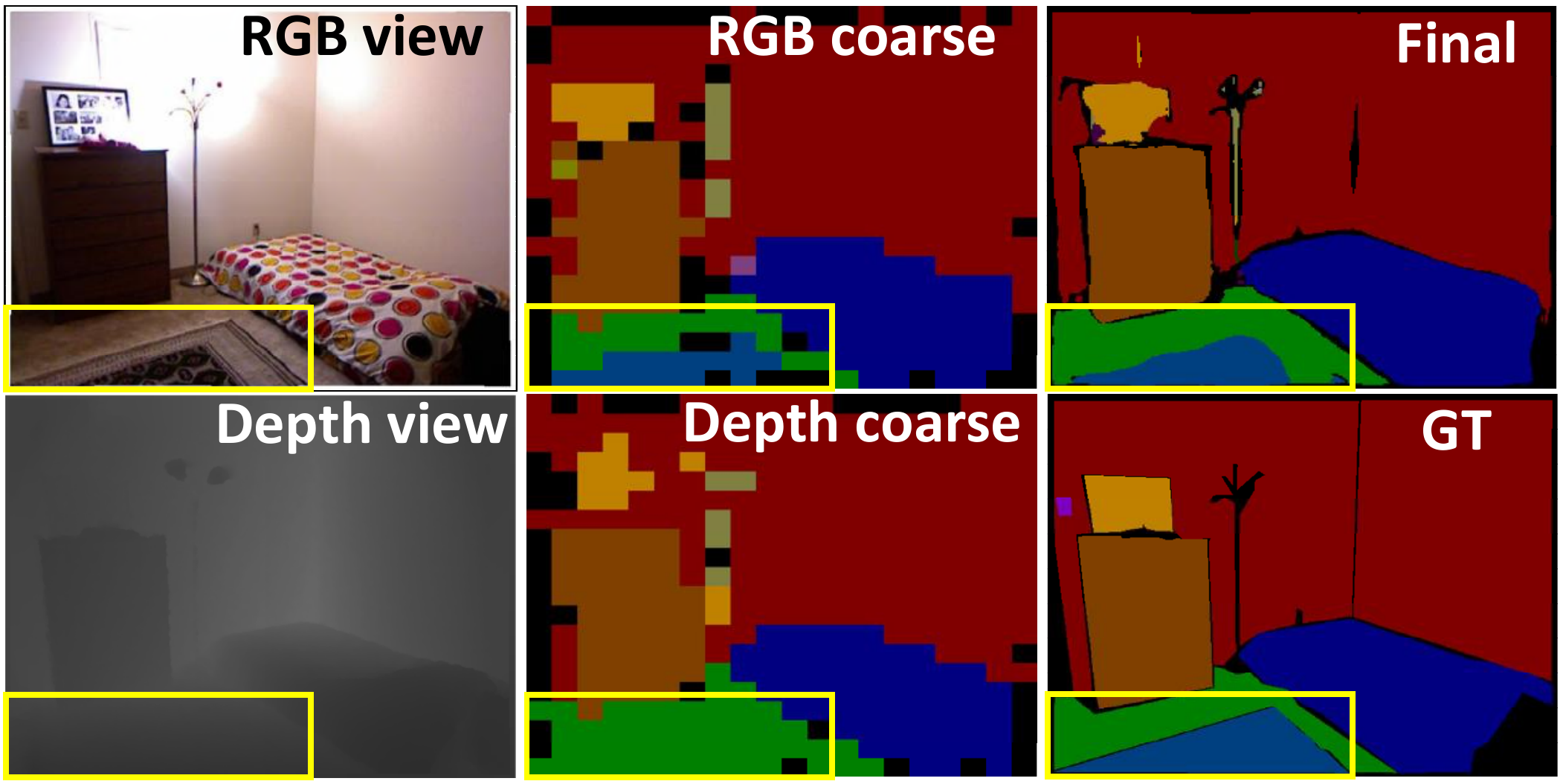}
\vspace{-5pt}
\caption{\fdp{An testing example of applying \ourmodel~to RGB-D semantic segmentation. Although the class ``floormat'' (yellow box) is almost indistinguishable in the depth view, its main part is correctly identified in the final prediction.}}
\label{fig_semantic}
\end{figure}

\subsection{\fkr{Linking to RGB-D Semantic Segmentation}}\label{sec47}
\fdp{To the best of our knowledge, almost no an existing model has adopted the Siamese network for RGB-D semantic segmentation. In contrast, most of them adopt a two-stream middle-fusion fashion \cite{hazirbas2016fusenet,wang2016learning,park2017rdfnet,deng2019rfbnet,chen2020bi}. The proposed \ourmodel~can be adapted to this task by simply replacing the prediction heads \cite{li2016deepsaliency}, \emph{i.e.,} changing the two $(1 \times 1,1)$ convolutional layers before coarse/final prediction into $(1\times 1,\mathcal{C})$ convolutions, where $\mathcal{C}$ indicates the number of classes in the semantic segmentation. Then, the network is trained with a pixel-wise multi-class cross-entropy loss against the ground truth label map. Following the standard 40-class train/test protocol on NYUDv2 \cite{gupta2014learning,chen2020bi,park2017rdfnet}, we obtained 35.0\% mIOU by directly applying \ourmodel~(Fig. \ref{fig_semantic}) without any other modification, and such a result is shown feasible on this task according to \cite{Long2017Fully}.}

\fdp{
We note that a potential challenge of using the Siamese network for RGB-D semantic segmentation is the huge commonality gap between RGB and depth modalities on this task, because RGB-D semantic segmentation is to identify \emph{category-specified} regions, where RGB and depth present huge gaps (\ie, weak commonality). This is in clear contrast to RGB-D SOD, where \emph{category-agnostic} salient objects usually ``pop out'' consistently in the two modalities, as illustrated in Fig. \ref{fig_motivation} and Fig. \ref{fig_multimoda}.  This actually motivates a topic that, besides RGB-D SOD and the applications shown in this paper, in what other cases a Siamese network is suitable. We believe such a topic is interesting and worthy of deeper investigation in the future.
}

\fkr{To better understand how \ourmodel~performs against existing semantic segmentation models and bridge these two fields, we carefully adapt several open-source SOTA segmentation models including PSPNet \cite{zhao2017pyramid}, RDFNet \cite{park2017rdfnet}, DANet \cite{fu2019dual}, SA-Gate \cite{chen2020bi}, and SGNet \cite{Chen2021spatial}, to the RGB-D SOD task.
We replace their multi-class classification heads with the corresponding saliency prediction heads (as mentioned above), and conduct evaluation on RGB-D SOD datasets. Note that RDFNet \cite{park2017rdfnet}, SA-Gate \cite{chen2020bi}, and SGNet \cite{Chen2021spatial} are three RGB-D semantic segmentation models, and they can be transferred directly. While PSPNet \cite{zhao2017pyramid} and DANet \cite{fu2019dual} are two representative RGB segmentation models, we adapt them by the late-fusion strategy adopted in \cite{Long2017Fully}. Also, HHA maps \cite{gupta2014learning,park2017rdfnet,chen2020bi} originally used by some RGB-D semantic segmentation models like RDFNet and SA-Gate were replaced with three-channel depth maps as input in our experiments for fair comparison. Comparative results are shown in Table \ref{tab_semantic}, where all the models were based on ResNet-101 and re-trained using the same training dataset as \ourmodel. We can see that \ourmodel~generally outperforms these semantic segmentation models on five representative datasets. Interestingly, we have also observed ``not bad'' results from some SOTA models, especially the latest SA-Gate, which even performs better than those tailored models in Table \ref{tab_sota}. This fact, experimentally reveals the underlying connection and transferability between these two fields, which we believe is interesting to study in the future. Besides, their differences may also exist, as indicated by the less satisfactory behavior of SGNet. The degraded performance of SGNet on this task is probably caused by the fact it relies on depth as a guidance to filter RGB features. However, in the task of RGB-D SOD, depth information may become less reliable. Another issue regarding these models we have observed is their coarse prediction with large output strides, leading to less accurate boundary details.}

\begin{table}[!htb]
\renewcommand{\arraystretch}{0.8}
\caption{Comparing \ourmodel~to existing semantic segmentation models transferred to the RGB-D saliency task. Symbol ``$\dag$'' means those RGB semantic segmentation models adapted to this task by the late-fusion strategy \cite{Long2017Fully}.
}\label{tab_semantic}
\centering
\footnotesize
{
\setlength{\tabcolsep}{0.4mm}
\begin{tabular}{r|cc|cc|cc|cc|cc}

\hline
             & \multicolumn{2}{c|}{\textit{NJU2K}}  & \multicolumn{2}{c|}{\textit{NLPR}}  & \multicolumn{2}{c|}{\textit{STERE}}  & \multicolumn{2}{c|}{\textit{RGBD135}}  & \multicolumn{2}{c}{\textit{SIP}} \\

             & \multicolumn{2}{c|}{\cite{ju2014depth}}  & \multicolumn{2}{c|}{\cite{peng2014rgbd}}  & \multicolumn{2}{c|}{\cite{niu2012leveraging}}  & \multicolumn{2}{c|}{\cite{cheng2014depth}}  & \multicolumn{2}{c}{\cite{fan2019rethinking}} \\
\cline{2-11}
             Model    & $S_{\alpha}\uparrow$ & $M\downarrow$   & $S_{\alpha}\uparrow$ & $M\downarrow$  & $S_{\alpha}\uparrow$ & $M\downarrow$  & $S_{\alpha}\uparrow$ & $M\downarrow$  & $S_{\alpha}\uparrow$ & $M\downarrow$ \\
\hline
\hline
\scriptsize{PSPNet$^\dag$\cite{zhao2017pyramid}}       & 0.901  & 0.045 & 0.918 & 0.028  & 0.899 & 0.046 & 0.909  & 0.026 & 0.856 & 0.066  \\
\scriptsize{RDFNet \cite{park2017rdfnet}}       & 0.891  & 0.050 & 0.910 & 0.031  & 0.897 & 0.047 & 0.919  & 0.027 & 0.875 & 0.055  \\
\scriptsize{DANet$^\dag$\cite{fu2019dual}}       & 0.900  & 0.044 & 0.912 & 0.027  & 0.889 & 0.048 & 0.896  & 0.027 & 0.870 & 0.056  \\
\scriptsize{SA-Gate \cite{chen2020bi}}       & 0.898  & 0.051 & 0.923 & 0.028  & 0.896 & 0.054 & \textbf{0.941}  & 0.022 & 0.874 & 0.059  \\
\scriptsize{SGNet \cite{Chen2021spatial}}       & 0.873  & 0.060 & 0.888 & 0.039  & 0.883 & 0.055 & 0.899  & 0.034 & 0.832 & 0.075  \\
\hline
\ourmodel &  \textbf{0.903}  & \textbf{0.043} & \textbf{0.925} & \textbf{0.022}  & \textbf{0.905} & \textbf{0.042} & 0.929  & \textbf{0.022} & \textbf{0.879} & \textbf{0.051}  \\
\hline
\end{tabular}
}\vspace{-0.2cm}
\end{table}
\section{Conclusion}\label{sec5}

We present a novel framework for RGB-D based SOD, named \ourmodel, which is based on joint learning and densely cooperative fusion.
Experimental results show the feasibility of learning a Siamese network for salient object localization in RGB and depth views, simultaneously, to achieve accurate prediction. Moreover, the densely cooperative fusion strategy employed is effective for exploiting cross-modal complementarity. \ourmodel~shows superior performance against SOTAs on seven benchmark datasets and is supported by comprehensive ablation studies. The generality and robustness of our framework has also been validated on two closely related tasks, \ie, RGB-Thermal (RGB-T) SOD and VSOD, \fkr{and also by comparing with SOTA semantic segmentation models}. The SOTA performance of \ourmodel~shows it could become a unified framework for multi-modal feature learning and fusion tasks, and we hope this work would serve as a catalyst for progressing many cross-modal tasks in the future.

\noindent \textbf{Acknowledgments.}\quad
This work was supported in part by the NSFC, under No. 61703077, 61773270, 61971005, U19A2052. We thank Yao Jiang and Suhang Li for their help on implementing \ourmodel~in Pytorch.

{
\bibliographystyle{IEEEtran}
\bibliography{rgbd_egbib}
}

\end{document}